\documentclass[twoside,11pt]{article}

\usepackage{jmlr2e}

\usepackage{xcolor}
\usepackage{bbm}
\usepackage{mathtools}

\firstpageno{1}

\newcommand{\yf}[1]{{#1}}

\usepackage{amsmath,amsfonts,bm}
\usepackage{amssymb}
\usepackage{bbm}

\def\eqref#1{equation~\ref{#1}}

\def\1{\bm{1}}
\newcommand{\train}{\mathcal{D}}

\def\vs{{\bm{s}}}

\DeclareMathAlphabet{\mathsfit}{\encodingdefault}{\sfdefault}{m}{sl}
\SetMathAlphabet{\mathsfit}{bold}{\encodingdefault}{\sfdefault}{bx}{n}

\def\gA{{\mathcal{A}}}

\def\gD{{\mathcal{D}}}

\def\gF{{\mathcal{F}}}
\def\gG{{\mathcal{G}}}

\def\gL{{\mathcal{L}}}

\def\gO{{\mathcal{O}}}

\def\gQ{{\mathcal{Q}}}
\def\gR{{\mathcal{R}}}

\def\gT{{\mathcal{T}}}
\def\gU{{\mathcal{U}}}

\def\sR{{\mathbb{R}}}
\def\sS{{\mathbb{S}}}

\newcommand{\E}{\mathbb{E}}

\newcommand{\Var}{\mathrm{Var}}

\DeclareMathOperator*{\argmax}{arg\,max}

\DeclareMathOperator*{\var}{Var}

\def\ie{\textit{i.e.,~}}
\def\eg{\textit{e.g.,~}}

\def\vs{\textit{v.s.~}}

\newcommand{\indep}{\perp \!\!\! \perp}

\usepackage{url}

\newtheorem{Theorem}[theorem]{Theorem}
\newtheorem{Lemma}[theorem]{Lemma}
\newtheorem{prop}[theorem]{Proposition}
\newtheorem{Definition}[theorem]{Definition}
\newtheorem{assumption}[theorem]{Assumption}
\newtheorem{Corollary}[theorem]{Corollary}

\usepackage{enumitem}
\usepackage{wrapfig}

\usepackage{enumitem}
\usepackage{wrapfig}

\usepackage{algorithm} 
\usepackage{algpseudocode} 
\usepackage{booktabs} %
\usepackage{threeparttable}  %
\usepackage{multirow} %
\usepackage{graphicx}
\usepackage{float}
\usepackage{color}
\usepackage{enumitem}
\usepackage[hang]{subfigure}

\usepackage{hyperref}

\usepackage{lastpage}
\jmlrheading{26}{2025}{1-\pageref{LastPage}}{9/22; Revised
6/25}{10/25}{22-1009}{Qi Zhang, Yifei Wang, and Yisen Wang}
\ShortHeadings{An Augmentation Overlap Theory of Contrastive Learning}{Zhang, Wang, and Wang}

\begin{document}

\title{An Augmentation
Overlap Theory of Contrastive Learning}
\author{\name Qi Zhang $^1$ $^*$
\email zhangq327@stu.pku.edu.cn 
\AND 
\name Yifei Wang $^2$ \thanks{Equal Contribution}
\email yifei\_w@mit.edu
\AND
\name Yisen Wang $^{1,3}$ \thanks{Corresponding Author}
\email yisen.wang@pku.edu.cn \\~\\
\addr $^1$State Key Lab of General Artificial Intelligence, School of Intelligence Science and Technology, Peking University \\
\addr $^2$CSAIL, MIT \\
\addr $^3$Institute for Artificial Intelligence, Peking University}
\editor{Pierre Alquier}
\maketitle

\begin{abstract}%
Recently, self-supervised contrastive learning has achieved great success on various tasks. However, its underlying working mechanism is yet unclear. In this paper, we first provide the tightest bounds based on the widely adopted assumption of conditional independence. Further, we relax the conditional independence assumption to a more practical assumption of augmentation overlap and derive the asymptotically closed bounds for the downstream performance. Our proposed augmentation overlap theory hinges on the insight that the support of different intra-class samples will become more overlapped under aggressive data augmentations, thus simply aligning the positive samples (augmented views of the same sample) could make contrastive learning cluster intra-class samples together. Moreover, from the newly derived augmentation overlap perspective, we develop an unsupervised metric for the representation evaluation of contrastive learning, which aligns well with the downstream performance almost without relying on additional modules. Code is available at https://github.com/PKU-ML/GARC.

\end{abstract}


\section{Introduction}
\label{sec:introduction}

\begin{figure}[!t]
    \centering
   \includegraphics[width=0.8\textwidth]{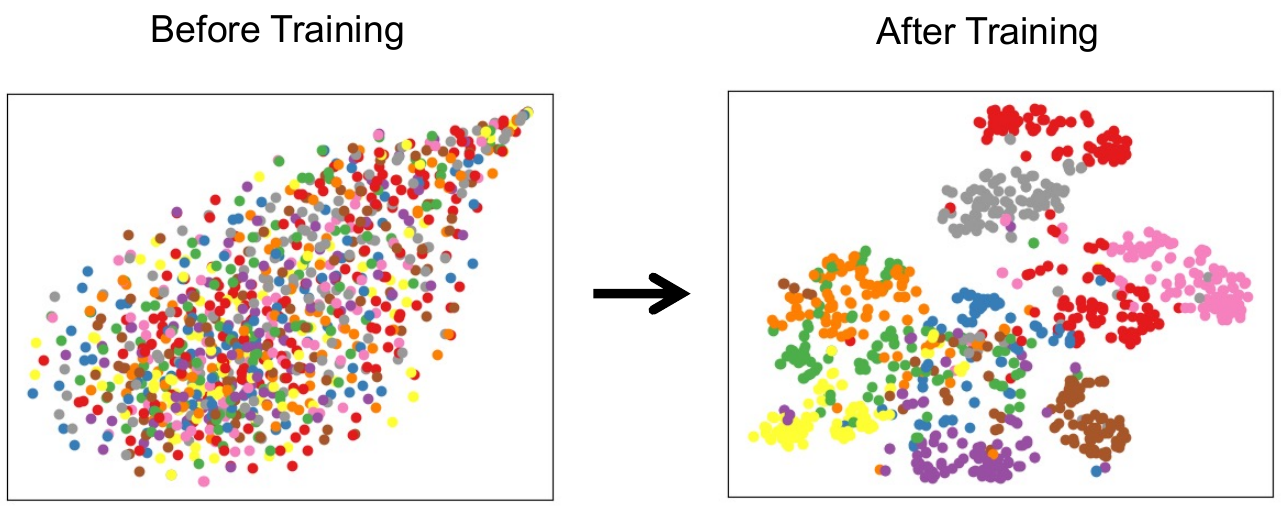}
    \caption{The t-SNE visualization of representations before and after contrastive learning method of SimCLR on CIFAR-10 data set. Each point denotes a sample and its color denotes its class.}
    \label{fig:clusted-features}
\end{figure}

{
Traditionally, supervised learning obtains representations by pulling together samples with the same labels (positive pairs) and pushing apart those with different labels (negative pairs). To improve the effectiveness of representation learning, early works found that the selection of positive and negative pairs plays a crucial role. This led to the development of algorithms such as triplet mining \citep{schroff2015facenet} and hard negative sampling \citep{bucher2016hard}, which aim to select more informative or challenging pairs from labeled data sets. However, obtaining such labels at scale can be costly. To remove this dependence on labeled data, recent research has focused on identifying surrogate signals for semantic similarity in an unsupervised manner. As a solution, self-supervised contrastive learning \citep{InfoNCE} generates positive pairs by applying two independent augmentations to the same input sample, under the assumption that augmented views should preserve semantic identity. Negative samples are typically drawn from other instances in the batch, assuming they are semantically different. When the number of classes is large, the chance of sampling a semantically similar (false negative) instance is low (approximately $1/K$ for $K$ classes), which makes this assumption practical. This technique enables the model to learn class-separated representations without access to labels, as shown in Figure \ref{fig:clusted-features}, and has become a cornerstone of self-supervised learning. Modern frameworks \citep{simclr,moco,BYOL,wang2021residual,oquab2023dinov2,wang2023message} have further refined this idea and {now achieved} performance comparable to supervised learning across a range of tasks.
} 
However, it is still unclear why contrastive learning can learn a meaningful representation for the downstream tasks and a convincing theoretical analysis is yet wanted.

\begin{figure}
    \centering
    \subfigure[Frameworks]{
    \label{fig:cl framework}
    \includegraphics[width=0.6\textwidth]{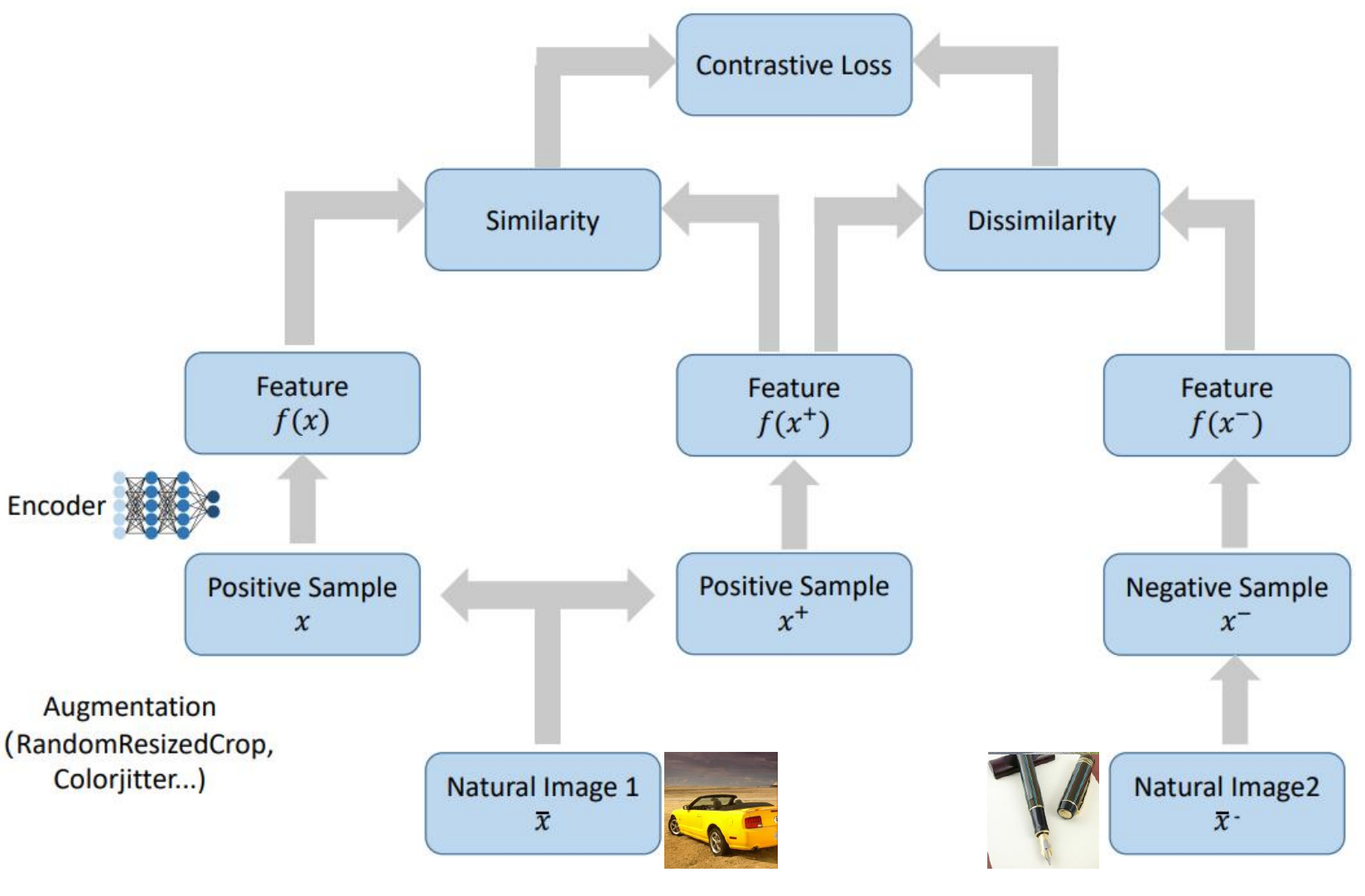}}
    \hfill
    \subfigure[Augmented views]{
    \label{fig:positive-samples}
    \includegraphics[width=0.3\textwidth]{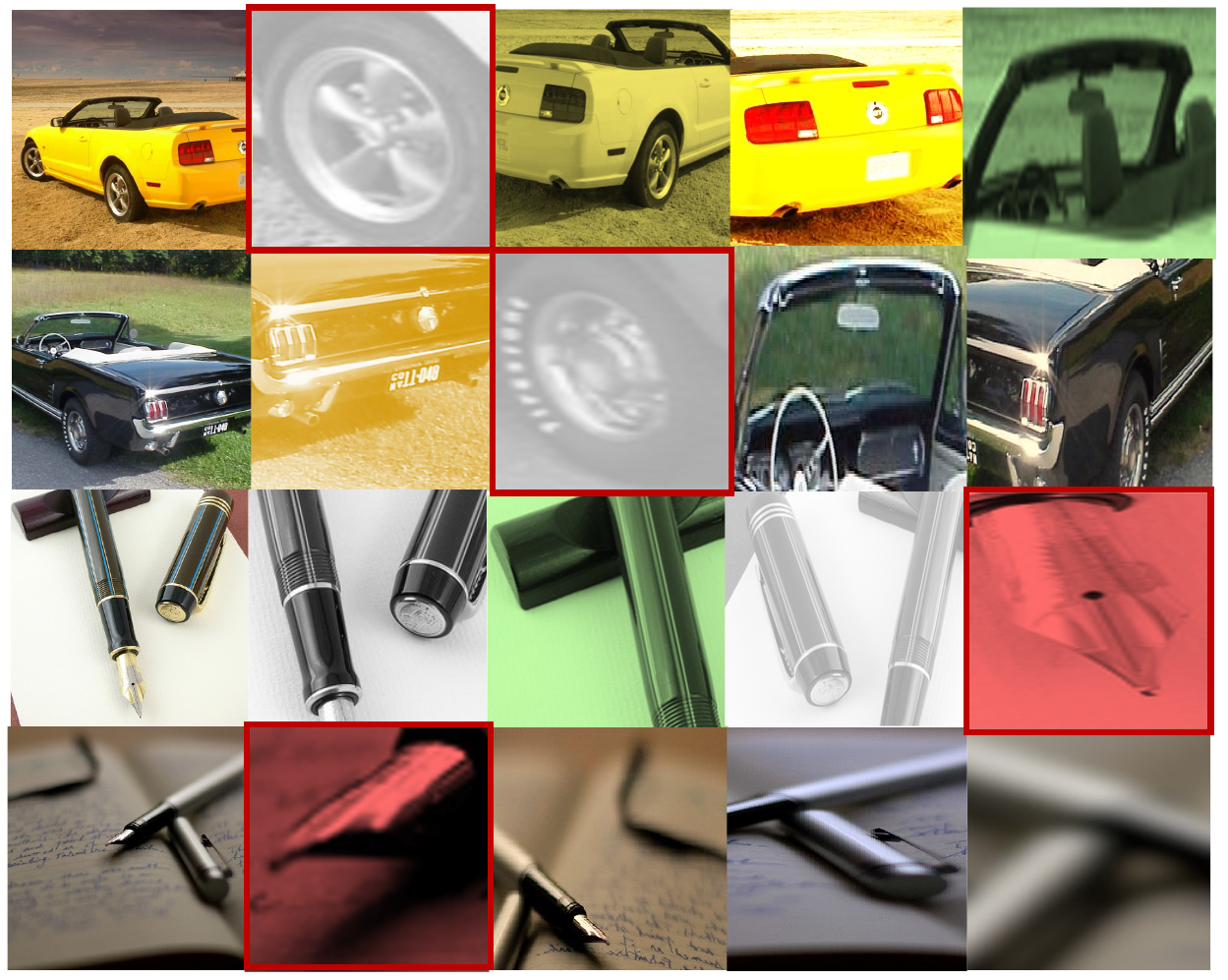}}
    \vspace{-0.1in}
    \caption{(a) The framework of contrastive learning. (b) The augmented views of four images from ImageNet. The first two rows are cars while the bottom two rows are pens. The ancher samples are shown in the 1st column while the 2-5th columns present its corresponding augmented views.}
    \label{fig:my_label}
    \vspace{-0.1in}
\end{figure}

        The core idea of contrastive learning is to learn an encoder by closing the augmented views of the same anchor sample while simultaneously pushing away the augmented views of different anchor samples in the feature space. { As shown in Figure \ref{fig:cl framework}, taking an image sample $\bar{x}$ as an example, contrastive learning first transforms it with data augmentations $t,t^+$ (e.g., RandomResizedCrop, ColorJitter, GaussianBlur, etc.) to generate a positive pair $(x,x^+)$. Then, the two augmented samples are respectively encoded with the encoder $f$. In the next step, contrastive learning trains the encoder with the contrastive loss (e.g., InfoNCE loss \citep{InfoNCE}) to decrease the distance between positive samples in the feature space and push away other samples.}
{ Intuitively, while supervised learning clusters samples based on their labels, contrastive learning performs instance-level discrimination without access to label information. Nevertheless, as shown in Figure \ref{fig:clusted-features}, when we compare the feature distributions before and after contrastive learning, we observe that the learned representations exhibit clear class separation. That is, samples from the same class are grouped together, whereas those from different classes are pushed apart—even without using any label supervision.} Therefore, it is natural to raise several questions: 
\begin{quote}
   \emph{Why the instance-level contrastive learning can lead to good performance on class-level downstream tasks? How to establish the theoretical relationship between (contrastive learning) pretraining and downstream performance?}
\end{quote}

\citet{arora} established a theoretical guarantee between (contrastive learning) pretraining and downstream tasks based on the assumption that the augmented views of the same sample are conditionally independent on its label. However, as shown in Figure \ref{fig:positive-samples}, the augmented views of the same sample are input-dependent and their assumption is quite hard to reach in practice. {Parallel to that, \citet{wang} decomposed the objective of contrastive learning to two parts: the alignment of positive samples and the uniformity of negative samples. While we find that only the alignment and the uniformity are not enough to make contrastive learning learn useful representations for the downstream tasks. In this paper, we propose a counter-example (Proposition 5.3) by constructing special positive pairs and prove that the downstream performance can be as bad as a random guess even when the learned representations achieve perfect alignment and uniformity. It means that except for the alignment and uniformity, the selection of positive pairs, i.e., the design of data augmentations, is crucial for contrastive learning, which may be overlooked by the above two kinds of works. Revisiting Figure \ref{fig:positive-samples}, we find that appropriate data augmentations can help intra-class samples generate quite similar augmented views (tires of different cars) while keeping the augmented views of inter-class samples distinguishable (cars and pens), which means that intra-class samples own the support overlap through these similar augmented views generated by data augmentations. As a result, when contrastive learning aligns the positive pairs, the intra-class samples that have overlapped views will be gathered as well.}

Motivated by the above observations, we formulate the data augmentations as an augmentation graph to bridge different intra-class samples. Based on which, we further propose a theory of augmentation overlap to analyze how the intra-class samples are eventually aligned by these overlapped views in contrastive learning. By characterizing the property of the augmentation overlap, we establish sharper upper and lower bounds for the downstream performance of contrastive learning under more practical assumptions. Moreover, we design a theory guided model selection metric via measuring the degree of augmentation overlap, which is totally unsupervised without additional computational costs. The contents of the paper are organized as follows:
\begin{itemize}
    \item In Section \ref{sec:related-work}, we summarize the related work of understanding self-supervised learning 
    and further introduce the problem setup in Section \ref{sec:problem setup}. 
    
    \item \yf{In Section \ref{sec:sharper bound with CI}, we derive sharper bounds for the downstream performance of contrastive learning by analyzing the influence of negative samples. To focus on negative samples, we adopt the widely used conditional independence assumption on positive samples and show how to improve the error bound on negative samples from {the Monte Carlo perspective}.} Our bounds are the tightest by far.
    
    \item \yf{In Section \ref{sec:beyond CI}, we focus on the positive samples and discuss how to relax the conditional independence assumption to a more realistic augmentation overlap assumption}, and propose a new theory of augmentation overlap to derive new upper and lower bounds for contrastive learning. 
    
    \item 
    In Section \ref{sec:effect of augmentation graph}, we present a quantitative analysis on the effect of data augmentation strategy on the augmentation overlap both theoretically and empirically. {In brief, strong augmentation strengths are necessary for bridging intra-class samples, which facilitates good downstream performance. However, too strong augmentations can lead to detrimental effects, such as the overlap of views between different classes.}
    \item Lastly, in Section \ref{sec:ARC total}, as a proof-of-idea, we demonstrate a practical application of our augmentation overlap theory, that is, we develop a representation evaluation metric for automatic model selection without using supervised downstream data.
\end{itemize}

\textbf{\textit{Remark.}} A preliminary work was published at ICLR 2022 \citep{wang2022chaos}, while we add substantially new results, both theoretically and empirically, in this manuscript, aiming to provide a more comprehensive and in-depth analysis of the generalization property of contrastive learning. Specifically, we add the following key results:
\begin{itemize}
    \item In Section \ref{sec:sharper bound with CI}, we establish \textit{sharper generalization bounds} under the conventional conditional independence assumption, which are superior to prior work and near-optimal. 
    \item In Section \ref{sec:beyond CI}, we generalize our prior bounds under perfect alignment to \textit{much weaker (more practical) assumptions}, and establish its relationship to the spectral properties of the augmentation graph.
    \item In Section \ref{sec:effect of augmentation graph}, we add a new theoretical discussion on the effect of data augmentation strength based on \textit{random graphs}, as well as empirically verifying these properties on real-world data sets.
    \item In Section \ref{sec:garc}, we extend our preliminary study on the model selection metric by studying its \textit{various variants}, and evaluate its effectiveness on real-world scenarios by comparison with state-of-the-art methods.
\end{itemize}

\section{Related Work}
\label{sec:related-work}

Self-supervised learning learns representations by designing appropriate surrogate tasks, like rotation prediction \citep{komodakis2018unsupervised,wang2021residual}, masked image reconstruction \citep{he2022masked,zhang2022mask,du2024on}, and instance discrimination \citep{moco,zhang2023on}. Among the self-supervised methods, contrastive learning is a representative type of framework which learns an encoder by closing the feature distance of positive samples generated by data augmentations \citep{simclr,BYOL,moco,zbontar2021barlow,caron2021emerging,oquab2023dinov2,cui2023rethinking}. Contrastive learning is quickly approaching the performance of supervised learning on different tasks and large-scale data sets (\eg ImageNet). Despite its huge success on empirical performance, the working mechanism of contrastive learning is still under explorations. In the following, we briefly summarize the related work from two views.

\textbf{\textit{1) View of Pretraining Objectives.}} InfoNCE \citep{InfoNCE} is a commonly used loss function in contrastive learning, which is motivated by NCE \citep{gutmann2012noise} to learn meaningful representations by measuring the mutual information of positive samples. However, some previous works show that the surrogate mutual information estimation of positive samples can not directly ensure the downstream performance of contrastive learning \citep{kolesnikov2019revisiting,Tschannen2020On}. There are other works analyzing contrastive learning from different perspectives. For example, \citet{wang} think contrastive learning is composed of two properties: alignment of positive samples and uniformity of negative samples. \citet{wang2021understanding} regarded InfoNCE as a hardness-aware function and analyzed the important temperature parameter $\tau$. \citet{wen2021toward} theoretically proved that contrastive learning can obtain sparse feature representations with appropriate data augmentations. \citet{zimmermann2021contrastive} indicated learned representations by contrastive learning implicitly invert the data generation process. \cite{hu2022your} reveal the connection between contrastive learning and stochastic neighbor
 embedding while \citet{wang2023message} establish the connection between contrastive learning and graph neural networks. \cite{parulekar2023infonce} theoretically prove that the representations learned by InfoNCE is well-clustered
 by the semantics in the data. Besides, some self-supervised learning paradigms without negative samples also achieve impressive performance \citep{BYOL}. Among them, \citet{tian2021understanding} theoretically proved that these types of methods will not collapse into trivial representations with the help of their proposed ingredients including Exponential Moving Average (EMA) and stop-gradient, while \citet{zhuo2023towards} analyzed it from the perspective of rank differential mechanism.

\textbf{\textit{2) View of Downstream Generalization.}}
\cite{arora} and \citet{jason} theoretically proved that the downstream performance of contrastive learning is close to that of supervised learning and the difference is controlled by an analyzable error term. However, their conclusions are based on an impractical assumption that the positive samples are conditionally independent{\footnote{Analysis on its impracticality could be found in Appendix \ref{sec:unpractice_of_CI}.}}, and the error term will get larger when contrastive learning has more negative samples (this is in contrast to empirical results).  
Following them, \cite{nozawa2021understanding} provided a new kind of theoretical analysis without conditionally independent assumption. However, there exists a class collision term in their bounds that can not be ignored. \cite{lei2023generalization} further establish generalization guarantees that do not rely on the number of negative examples.
Besides, there are some other perspectives to theoretically characterize the downstream performance of contrastive learning, including graph theory \citep{haochen2021provable,wang2024do}, information theory {\citep{InfoMin,tsai2020self,tosh2020contrastive,ouyang2025projection,cui2025an}}, kernel method \citep{li2021self}, causal mechanism \citep{mitrovic2021representation}, and distributionally robust optimization \citep{wu2023understanding}. 

\textbf{\textit{Other Types of Self-Supervised Learning Methods.}} Beside contrastive learning, there are other types of self-supervised learning methods. For reconstruction-based methods, 
\citet{garg2020functional} established theoretical analysis by viewing them as imposing a regularization on the
representation via a learnable function, and \citet{cao2022understand,zhang2022mask} theoretically analyzed the function of different designs for the reconsturction-based method  \citep{he2022masked}. \cite{wang2024rethinking} understands reconstruction-based methods from two properties following \cite{wang}, i.e., the alignment and uniformity. \cite{tan2023information} analyze the reconsturtcion-based objectives from the matrix information theory. For auto-regressive based methods, \citet{saunshi2021a} reformulated text classification tasks as sentence completion problems and proposed theoretical guarantees for their downstream performance. \cite{zhang2024look} theoretically compare reconstruction and autoregressive-based methods from a spectral perspective.

\section{Problem Setup}
\label{sec:problem setup}
Here, we briefly introduce some basic notations and common practice of contrastive learning for the image classification task. Generally, it has two stages, unsupervised pretraining and supervised finetuning. 
{
 In the first stage, on the unlabeled data $\train_u=\{\bar x\}$, we pretrain an encoder $f\in\gF$, where $\gF$ is a hypothesis class consists of functions mapping from the $d$-dimensional input space $\sR^d$ to the unit hypersphere $\sS^{m-1}$ in $m$-dimensional space. We assume that $f$ is normalized, meaning the representation space $\sS^{m-1}$ is bounded. This assumption is consistent with the designs of common contrastive learning frameworks like SimCLR \citep{simclr} and MoCo \citep{moco}.
 }
In the second stage, we evaluate the learned representations $z$ with the labeled data $\train_l=\{(\bar x,y)\}$ where label $y\in \{1,2,\dots,K\}$. For simplicity, we assume that every sample belongs to a unique class, \ie $p(y|x)$ being one-hot, and the data set is class balanced, \ie $p(y=k)=1/K$.

\textbf{\textit{Contrastive Pretraining.}}
Given a random raw training example $\bar x\in\train_u$, we first draw two \textit{positive samples} $(x, x^+)$ by applying two randomly drawn data augmentations $t,t^+\in\gT$ to $\bar x$, \ie $x=t(\bar x), x^+=t^+(\bar x)$, where $\gT=\{t:\sR^d\to\sR^d\}$ contains all possible predefined data augmentations. Let $p(x,x^+)$ be the joint distribution of positive pairs, and $p(x)=\int p(x,x^+)dx^+$ be the marginal distribution of the augmented data $x$. 
Without loss of generality, we take $x$ as the anchor sample, and draw $M$ augmented samples $\{x_i^-\}_{i=1}^M$ independently from the marginal distribution $p(\cdot)$ as its \textit{negative samples}, and denote their corresponding distributions as $p(x^-_i),i=1,\dots,M$.
Then, the encoder $f$ is learned by the contrastive loss. { Researchers have designed various types of contrastive loss like triplet loss \citep{schroff2015facenet}, hinge loss \citep{al2020robust}, margin loss \citep{shah2022max}, etc.}
Following the same spirit (pulling in positive pairs and pushing away negative pairs), we mainly consider a most widely used InfoNCE loss \citep{InfoNCE} in this paper:
\begin{equation}
\begin{aligned}
\gL_{\rm contr}(f)
=&\E_{p(x,x^+)\Pi_ip(x_i^-)}\left[-\log\frac{\exp(f(x)^\top f(x^+))}{\sum_{i=1}^M\exp(f(x)^\top f(x^-_i))}\right].\\
\end{aligned}
\label{eqn:infonce-frac}
\end{equation}
In practice, some variants like SimCLR \citep{simclr} also include the positive pair $(x,x^+)$ in the denominator (inside the logarithm) of the InfoNCE loss. A recent work \citep{yeh2021decoupled}, however, notices that this extra term introduces a coupling between positive and negative samples, which leads to sample deficiency such as requiring a large batch size (\eg 4096). Instead, both \citet{yeh2021decoupled} and \citet{furst2021cloob} show that omitting the term of positive pairs leads to consistent performance improvement. Thus, we adopt this positive-free denominator in the paper.

\textbf{\textit{Evaluation on Downstream Tasks.}} To measure the quality of representations learned by contrastive pretraining, a popular metric is the prediction accuracy of a linear classifier $g(z)=[w^\top_1,\dots,w^\top_K]z$ learned on top of the learned representations $z=f(x)$.
Typically, the linear classifier is trained by the Cross Entropy (CE) loss \citep{simclr} on the labeled data $\gD_l$:
\begin{equation}
\gL_{\rm CE}(f)=\E_{p(x,y)}\left[
-\log\frac{\exp\left(f(x)^\top w_y\right)}{\sum_{i=1}^K\exp\left(f(x)^\top w_i\right)}\right],
\label{eq:linear-evaluation}
\end{equation}
{where $w_y$ is the optimal linear parameter for class $y$.} For the convenience of theoretical analysis, following the common practice in \cite{arora,ash2021investigating,nozawa2021understanding}, we directly adopt the mean representation of each class as the classwise weights, \ie $w_k=\mu_k:=\E_{p(x,y)}[f(x)|y=k]$, namely, \emph{mean classifier}. Thus, we obtain the following mean CE loss (mCE):
\begin{equation}
\gL_{\rm mCE}(f)=\E_{p(x,y)}\left[
-\log\frac{\exp\left(f(x)^\top \mu_y\right)}{\sum_{k=1}^K\exp\left(f(x)^\top \mu_k\right)}\right].
\label{eq:mean-linear-evaluation}
\end{equation}
It is straightforward to conclude that $\gL_{\rm CE}(f)\leq \gL_{\rm mCE}(f)$ and their performance is comparable in practice as demonstrated in \citet{arora}.

\textbf{\textit{Scale Adjustment.}}
For the theoretical analysis of contrastive learning, our goal is to characterize the gap between the pretraining loss $\gL_{\rm contr}(f)$ (Eq.~\ref{eqn:infonce-frac}) and the downstream classification loss $\gL_{\rm mCE}(f)$ (Eq.~\ref{eq:mean-linear-evaluation}). Revisiting Eq.~\ref{eqn:infonce-frac} and Eq.~\ref{eq:mean-linear-evaluation}, we find that their main difference lies on the denominator, \ie $\gL_{\rm contr}(f)$ is computed over $M$ negative samples while $\gL_{\rm mCE}(f)$ is over $K$ class centers.
The scale of $M$ and $K$ could be very different, which further leads to the scale mismatch between $\gL_{\rm contr}(f)$ and $\gL_{\rm mCE}(f)$. 
Therefore, we adjust the loss by taking the mean score instead of summation in the denominator accordingly: %

\begin{equation}
\bar{\gL}_{\rm contr}(f)
=\E_{p(x,x^+)\Pi_ip(x_i^-)}\left[-\log\frac{\exp(f(x)^\top f(x^+))}{{\frac{1}{M}}\sum_{i=1}^M\exp(f(x)^\top f(x^-_i))}\right]
={\gL}_{\rm contr}(f)+{\log(M^{-1})},
\label{eq:adjust-infonce}
\end{equation}

{
\begin{equation}
\bar{\gL}_{\rm mCE}(f)
=\E_{p(x,y)}\left[-\log\frac{\exp(f(x)^\top\mu_y)}{{\frac{1}{K}}\sum_{k=1}^K\exp(f(x)^\top \mu_k)}\right]
={\gL}_{\rm mCE}(f)+{\log(K^{-1})}.
\label{eq:adjust-mce}
\end{equation}
}
In this way, the two normalized objectives are irrelevant to the scale of $M$ and $K$. Note that the normalization has \textit{no effect} on the learning process as their gradients are equal to the original ones. In the following discussion, we mainly deal with the normalized versions. Additionally, regarding the discrepancy between the two objectives, the scale adjustment above amounts to adding a $\log(K/M)$ term to the contrastive loss:
\begin{equation}
\bar{\gL}_{\rm contr}(f)-\bar{\gL}_{\rm mCE}(f)={\gL}_{\rm contr}(f)+\log(K/M)-{\gL}_{\rm mCE}(f).
\end{equation}

\section{Improved Bounds under Conditional Independence Assumption}
\label{sec:sharper bound with CI}

{In this section, we focus on improving bounds by analyzing the roles of \textit{negative samples}. Without loss of generality, we take the \textit{positive pairs} under the conditional independence assumption \citep{arora} as an example, which enables us to have a fair comparison to previous bounds and shows our advantages more directly.
Specifically, we propose a new technique by showing the negative samples are closely related to the Monte Carlo estimation of downstream class centers, which provides a new insight on the benefits of more negative samples. Further, built upon this technique, we show that \textit{for the first time}, we could obtain an asymptotically closed upper bound on the downstream error. This new finding provides a solid guarantee on the downstream performance of contrastive learning and forms the basis for our further analysis beyond conditional independence in the next section.}

\begin{assumption}[Conditional Independence \citep{arora}]
\label{assumption:conditional-independence}
The two positive samples $x,x^+\sim p(x,x^+)$ are conditionally independent given the label $y$, \ie $x \indep x^+\mid y$.
\end{assumption}
In the following, we will show how the InfoNCE loss (Eq.~\ref{eq:adjust-infonce}) closely upper bounds the mean CE loss (Eq.~\ref{eq:adjust-mce})
under the conditional independence assumption. 
Firstly, we decouple the two objectives into a {positive objective and a negative objective}, respectively:
\begin{align}
\bar{\gL}_{\rm contr}(f)
&=\underbrace{-\E_{p(x,x^+)}[f(x)^\top f(x^+)]}_{\bar{\gL}_{\rm contr}^+(f)}+\underbrace{\E_{\Pi_ip(x^-_i)}\log\frac{1}{M}\sum_{i=1}^M\exp(f(x)^\top f(x^-_i))}_{\bar{\gL}_{\rm contr}^-(f)}, \\
\bar{\gL}_{\rm mCE}(f)&=\underbrace{-\E_{p(x,y)}[f(x)^\top\mu_y]}_{\bar{\gL}_{\rm mCE}^+(f)}{+\underbrace{\E_{p(x)}\log\left[\frac{1}{K}\sum_{k=1}^K\exp(f(x)^\top \mu_k)\right]}_{\bar{\gL}_{\rm mCE}^-(f)}}.
\end{align}
It is easy to see that the two positive objectives are equal under conditional independence:
\begin{equation}
\begin{aligned}
\bar{\gL}_{\rm contr}^+(f)&=-\E_{p(x,x^+)}[f(x)^\top f(x^+)]
=-\E_{p(y)p(x\mid y)p(x^+\mid y)}[f(x)^\top f(x^+)]\\
&=-\E_{p(y)p(x\mid y)}\left[f(x)^\top [\E_{p(x^+\mid y)} f(x^+)]\right]
=-\E_{p(y)p(x\mid y)}[f(x)^\top \mu_y]=\bar{\gL}_{\rm mCE}^+(f).
\end{aligned}
\label{eq:nce-pos}
\end{equation}
Thus, we only need to bound the two negative objectives ($\bar{\gL}_{\rm contr}^-(f)$ and $\bar{\gL}_{\rm mCE}^-(f)$). 

\textbf{\textit{Limitations of Previous Analysis.}}
\yf{Previous studies of the negative objective \citep{arora,nozawa2021understanding,ash2021investigating} often adopt a collision-coverage perspective to dissect the negative samples.} Specifically, they use the proportion of negative samples belonging to the positive class (class collision) and the probability that there are at least $K-1$ negative samples covering the $K-1$ negative classes (class coverage) to form an upper bound.
However, a clear drawback of this analysis is that it actually only requires $K$ negative samples for class coverage, and the other $M-K$ samples are discarded in the analysis, resulting in a very loose bound. In this work, we provide a new technique to analyze the role of $M$ negative samples. In particular, we show that all $M$ negative samples are useful and they all together contribute a tight upper bound that is asymptotically closed with $M\to\infty$. 
{
\citet{Bao2021OnTS} also adopt this technique to analyze contrastive learning with the { InfoNCE loss}. Nevertheless, their analysis involves applying a reverse Jensen's inequality \emph{twice}, leading to a large error in the upper bound that is not decreasing w.r.t.~$M$.}

\textbf{\textit{Our New View of Negative Samples for Monte Carlo Estimation.}} 
For the negative objective of InfoNCE loss, due to the convexity of the $\operatorname{logsumexp}$ operator, following Jensen's inequality we have 
\begin{equation}
\begin{aligned}
\bar{\gL}_{\rm contr}^-(f)&=\E_{\Pi_ip(y^-_i)p(x^-_i\mid y^-_i)}\log\frac{1}{M}\sum_{i=1}^M\exp(f(x)^\top f(x^-_i))\\
&\geq\E_{\Pi_ip(y^-_i)}\log\frac{1}{M}\sum_{i=1}^M\exp(\E_{p(x_i^-\mid y^-_i)}f(x)^\top f(x^-_i)) \\
&=\E_{\Pi_ip(y^-_i)}\log\frac{1}{M}\sum_{i=1}^M\exp(f(x)^\top\mu_{y^-_i}):=\bar{\gL}_{\rm MC}(f).
\end{aligned}
\label{eq:nce-mc}
\end{equation}
{As we assume $p(y=k) = 1/K$, it is easy to see that its lower bound $\bar{\gL}_{\rm MC}(f)$ is a (biased) Monte Carlo estimation of the negative objective of the mean CE loss, \ie $\bar{\gL}_{\rm mCE}^-(f)=\log\E_{p(y)}\exp(f(x)^\top\mu_{y})$, and the approximation error shrinks to 0 as $M\to\infty$, as characterized in the following lemma.} 
\begin{Lemma}
The approximation error of the Monte Carlo estimation $\bar{\gL}_{\rm MC}(f)$ shrinks in the order $\gO(M^{-1/2})$, specifically.
\begin{equation}
    |\bar{\gL}_{\rm MC}(f)-\bar{\gL}_{\rm mCE}^-(f)|\leq\frac{e}{\sqrt{M}}.
\label{eq:mc-mce}
\end{equation}
\label{lemma:mc-error}
\end{Lemma}
{
Lemma \ref{lemma:mc-error} shows that all $M$ negative samples contribute to a better Monte Carlo estimator of the negative objective of the mean CE loss.}

Combining Eqs.~\ref{eq:nce-pos}, \ref{eq:nce-mc} \& \ref{eq:mc-mce}, we derive an asymptotically closed upper bound on the downstream performance:
\begin{equation}
\begin{aligned}
\bar{\gL}_{\rm mCE}({f})&=\bar{\gL}^+_{\rm mCE}({f})+\bar{\gL}^-_{\rm mCE}({f})\leq\bar{\gL}^+_{\rm contr}({f})+\bar{\gL}_{\rm MC}({f})+\frac{e}{\sqrt{M}}\\
&\leq \bar{\gL}^+_{\rm contr}({f})+\bar{\gL}^-_{\rm contr}({f})+\frac{e}{\sqrt{M}}=\bar{\gL}_{\rm contr}({f})+\frac{e}{\sqrt{M}}.
\end{aligned}
\end{equation}
Following a similar routine, we can also derive the lower bound on the downstream performance via a reversed Jensen's inequality \citep{reverse_jensen}. 
Therefore, under the conditional independence assumption, our guarantees for the downstream performance of contrastive learning can be summarized as: 
\begin{Theorem}[Downstream Guarantees under Conditional Independence]
Under Assumption \ref{assumption:conditional-independence}, the downstream classification risk $\bar\gL_{\rm mCE}(f)$ of any $f \in \mathcal{F}$ can be upper and lower bounded by the contrastive risk $\bar\gL_{\rm contr}(f)$ as
\begin{equation}
\underbrace{\bar\gL_{\rm contr}(f)-\frac{1}{2}{\var(f(x)\mid y)}-\frac{e}{\sqrt{M}}}_{\text{lower bound } \gR_{\rm L}}\leq\bar\gL_{\rm mCE}({f})\leq\underbrace{\bar\gL_{\rm contr}({f}) + \frac{e}{\sqrt{M}}}_{\text{upper bound }\gR_{\rm U}},
\end{equation}
where $M$ is the number of negative samples. Furthermore, the conditional variance $\Var(f(x)|y)=\E_{p(x,y)}\Vert f(x)-\E_{p(\hat x|y)}f(\hat x)\Vert^2$ is at most $2$. As a result, the gap between the upper and the lower bounds can be bounded: $\gR_{\rm U}-\gR_{\rm L}\leq 1+2e/\sqrt{M}$.
\label{thm:bounds-CI}
\end{Theorem}

Theorem \ref{thm:bounds-CI} shows that the downstream performance can be upper and lower bounded by the contrastive loss. Thus, minimizing contrastive loss is almost equivalent (with a small surrogate gap) to optimizing the supervised loss, which helps us to understand the empirical effectiveness of contrastive learning. Furthermore, as $M\to\infty$, \ie adopting more negative samples, the upper and lower bounds can be further narrowed to ${\bar\gL_{\rm contr}(f)-\frac{1}{2}{\var(f(x)\mid y)}}\leq\bar\gL_{\rm mCE}({f})\leq{\bar\gL_{\rm contr}({f})}$. 
In the common practice of contrastive learning, a large number of negative samples indeed often brings better downstream performance \citep{simclr}. Our theoretical analysis not only echoes with this empirical finding, but also provides new insights on the role of negative samples. 
{According to Lemma 4.2, negative samples from all classes (even the same supervised class as positive samples) contribute to a better Monte Carlo estimation.
Thus, we can close the upper bound even in the presence of false negative samples.
This implication also aligns well with the practice, as contrastive learning indeed performs comparably or even superiorly to {supervised learning} on real-world data sets without pruning the false negative samples \citep{tomasev2022pushing}. While this does not preclude the potential benefits from negative sample mining (as explored in \cite{robinson2020contrastive}) in contrastive learning, as Lemma 4.2 only implies {the impact of the size} of negative samples for contrastive learning. As for the comparisons of the effects of contrastive learning with and without negative sample mining, it is outside the scope of Lemma 4.2.}

\begin{figure}[!htbp]
    \centering
    \subfigure[A randomly initialized encoder $f$.]{
    \includegraphics[width=.45\textwidth]{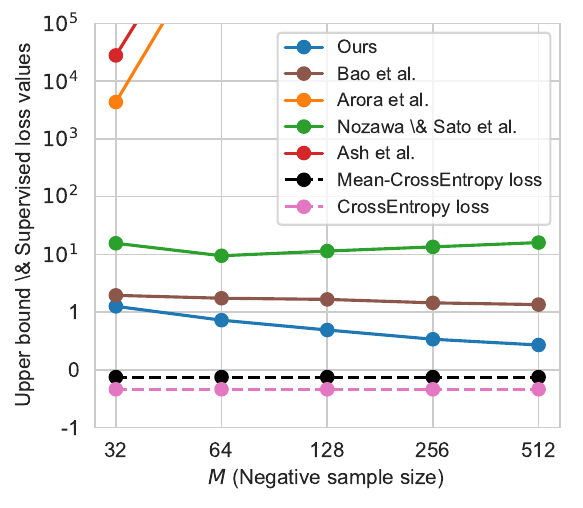}
    }
    \subfigure[An encoder $f$ trained with SimCLR.]{
    \includegraphics[width=.45\textwidth]{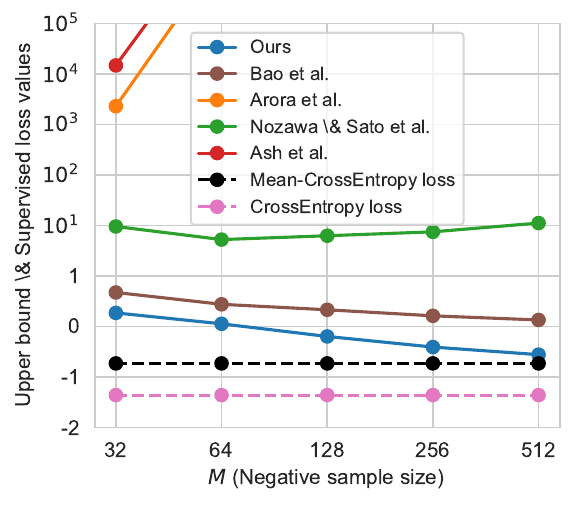}
    }
    \caption{Comparison of upper bounds on the downstream loss (measured by mean CE loss) on CIFAR-10. The encoder is a ResNet-18 \citep{he2016deep} and we train it using SimCLR \citep{simclr}. We calculate the upper bounds using its representations at the (a) initialization and (b) final stages.}
    \label{fig:different-bounds}
\end{figure}

\textbf{\textit{Discussion.}}
Prior to ours, several works \citep{arora,ash2021investigating,nozawa2021understanding,Bao2021OnTS} also propose different versions of guarantees for downstream performance, and we summarize their upper bounds in Table \ref{tab:upper-bounds}. As some works analyze other variants of the pretraining and downstream objectives, \eg hinge loss \citep{arora}, we generally denote them as $\gL_{\rm unsup}$ and $\gL_{\rm sup}$, respectively.
In particular, the seminal work of \citet{arora} draws two important implications from the upper bound: 1) the surrogate gap between the two objectives cannot be closed because of an unavoidable \textit{class collision} errors ($\tau_M,\mathrm{Col}$) in the upper bound that account for the proportion of negative samples belonging to the same class of the anchor sample, \ie the false negative samples; and 2) the overall upper bounds increase with more negative samples (larger $M$). However, both implications have been shown \textit{contradictory} to the empirical practice, as discussed above. 
Subsequential works are devoted to resolving the two issues by eliminating the class collision errors and demonstrating the benefits of a large $M$. Specifically, \cite{nozawa2021understanding}  and \cite{ash2021investigating} manage to show benefit from larger $M$, and \cite{Bao2021OnTS} fully eliminate the class collision error.
\yf{Nevertheless, their final upper bounds have unavoidable error terms even with $M\to\infty$.} Indeed, as shown in Figure \ref{fig:different-bounds}, as $M\to\infty$, the bounds of previous work are either explosive \citep{arora,ash2021investigating} or roughly constant \citep{nozawa2021understanding,Bao2021OnTS}. In comparison, with the proposed Monte Carlo analysis of negative samples, we take all negative samples into consideration and resolve the two issues completely. In particular, we are the first to show that the upper bound could be asymptotically tight with a large $M$, as shown theoretically in Table \ref{tab:upper-bounds} and empirically in Figure \ref{fig:different-bounds}. Our results suggest that contrastive learning is indeed \textit{almost equivalent} to supervised learning under the conditional independence assumption.

\begin{table}[!t]
\centering
\caption{A comparison of upper bounds of the supervised downstream loss $\gL_{\rm sup}(f)$ using the unsupervised pretraining loss $\gL_{\rm unsup}(f)$ (under the conditional independence assumption). Here, $\mathrm{Col}=\sum_{m=1}^M\mathbbm{1}[y_{x^-_m}=y_{x}]$ denotes the degree of class collision; $\tau_M$ denotes the collision probability that at least one of the $M$ negative samples belong to the positive class $y_x$; $v_{M+1}$ denotes the coverage probability that $M+1$ negative samples contains all classes $k\in[K]$; and $H_{K-1}=\sum_{k=1}^{K-1}1/k$ is the $(K-1)$-th harmonic number.
$^*$Adjusted objective scales for clear comparison.
}
\begin{tabular}{lll}
\toprule 
& Upper Bound & Reference \\
\midrule
\vspace{0.1in}
$\gL_{\rm sup}(f) \leq$ & $\frac{1}{\left(1-\tau_{M}\right) v_{M+1}}\Big(\gL_{\rm unsup}(f)-\mathbb{E} \log (\mathrm{Col}+1)\Big)$ & \cite{arora} \\
\vspace{0.1in}
& $\frac{1}{v_{M+1}}\left(2\gL_{\rm unsup}(f)-\mathbb{E} \log (\mathrm{Col}+1)\right)$ & \cite{nozawa2021understanding} \\
\vspace{0.1in}
& $\frac{2}{1-\tau_{M}}\left(\frac{2(K-1) H_{K-1}}{M}\right)\Big(\gL_{\rm unsup}(f)-\mathbb{E} \log (\mathrm{Col}+1)\Big)$ & \cite{ash2021investigating} \\
\vspace{0.1in}
& $\gL_{\rm unsup}(f) + 2\log(\cosh(1))$ & \citet{Bao2021OnTS}$^*$\\
& $\gL_{\rm unsup}(f) + e/\sqrt{M}$ & Our work$^*$ \\
\bottomrule
\end{tabular}
\label{tab:upper-bounds}
\end{table}

\section{From Conditional Independence to Augmentation Overlap}
\label{sec:beyond CI}

{ In Section 4, we have improved the bounds on negative samples from the Monte Carlo perspective}. While in this section, our focus shifts to elaborating on the effects of positive pairs. We first relax the impractical conditional independence assumption used in Theorem \ref{thm:bounds-CI} and derive a new bound that is unfortunately with an extra unbounded term of intra-class variance. To further analyze this variance term, we revisit contrastive learning from a graph perspective and build a new augmentation overlap framework. Based on that, we obtain a new guarantee for the gap between the pretraining and downstream performance of contrastive learning.

\subsection{Towards a Relaxation of Conditional Independence}

{ Intuitively, as shown in Figure 2, a positive pair is generated from the same natural image. Consequently, the positive pairs are input-dependent and it is hard to satisfy the conditional independence assumption in practice. }So we release the impractical conditional independence assumption and start with a basic assumption on the label consistency between positive samples, that is, any pair of positive samples $(x,x^+)$ should nearly belong to the same class. 
{
\begin{assumption}[Label Consistency]
$\forall\ x,x^+ \sim p(x,x^+)$, {we denote $y$ and $y^+$ as their labels. We assume the probability that they have different labels is smaller than $\alpha$, \ie 
\begin{equation}
\E_{x,x^+ \sim p(x,x^+)}\mathbbm{1}(y\neq y^+)\leq \alpha.  
\end{equation}
}
\label{assumption:label-consistency}
\end{assumption}
}
This is a basic and natural assumption that is likely to hold in practice. From Figure \ref{fig:positive-samples}, we can see that the widely adopted augmentations (\eg images cropping, color distortion, and horizontal flipping) will hardly alter the image belonging class.

{
\begin{Theorem}[Downstream Guarantees without Conditional Independence] If Assumption \ref{assumption:label-consistency} holds, then, for any $f \in \mathcal{F}$, its downstream classification risk $\bar\gL_{\rm mCE}(f)$ can be bounded by the contrastive learning risk $\bar\gL_{\rm contr}(f)$
\begin{equation}
\begin{aligned}
&\bar\gL_{\rm contr}(f)- {2\sqrt{\var(f(x)\mid y)}} - \frac{1}{2}{\var(f(x)\mid y)}-4\sqrt{\alpha}-\frac{e}{\sqrt{M}} \\
&\qquad \leq \bar\gL_{\rm mCE}({f})\leq \bar
\gL_{\rm contr}(f) + {2\sqrt{\var(f(x)|y)}} +4\sqrt{\alpha}+ \frac{e}{\sqrt{M}}.
\end{aligned}
\label{eq:general-gap}
\end{equation}
\label{thm:general-generalization-gap}
\end{Theorem}
}

Theorem \ref{thm:general-generalization-gap} shows that, even without the conditional independence assumption, we can still derive a bound for the downstream performance. Compared to Theorem \ref{thm:bounds-CI}, there is an additional intra-class variance term $\sqrt{\var(f(x)\mid y)}$ in the lower and upper bound. { Only with the label consistency assumption (Assumption \ref{assumption:label-consistency}), the variance term can not be bounded, which means that the generalization gap in Theorem \ref{thm:general-generalization-gap} could be quite large.}
For example, when the intra-class variance term is large enough, contrastive learning might have inferior performance as shown in Proposition \ref{prop:wang-couterexample}.

\begin{prop}
For $N$ training examples of $K$ classes, consider a case that inter-anchor features $\{f(x_i)\}_{i=1}^N$ are randomly distributed in $\sS^{m-1}$ while intra-anchor features are perfectly aligned, \ie $\forall x_i,x_i^+\sim p(x,x^+), f(x_i)=f(x^+_i)$. In this case, the expectation of the numerator of InfoNCE loss $\E_{ p(x_i,x_i^+)}(\exp(f(x_i)^\top f(x^+_i)))$ achieves its maximum while the expectation of the denominator $\E_{\Pi_ip(x_j^-)}(\frac{1}{M}\sum_{j=1}^M\exp(f(x_i)^\top f(x^-_j)))$ achieves its minimum, {i.e., both the alignment and uniformity losses \citep{wang} achieve the minimum,
thus the InfoNCE loss obtains its minimum}. However, the downstream classification accuracy is at most $1/K+\varepsilon$ where $\varepsilon$ is close to 0 when $N$ is large enough.
\label{prop:wang-couterexample}
\end{prop}
\vspace{-0.05in}

If given the conditional independence assumption, the alignment between positive samples is equivalent to the alignment between the sample itself and its class center. Thus, the intra-class samples will be finally aligned to the class center, which will not be uniformly distributed such that the above failure case does not exist. 

To summarize, the conditional independence assumption is too strong to eliminate the variance term in the theoretical bounds (Theorem \ref{thm:bounds-CI}), while discarding this assumption will make the variance term unbounded and further lose the guarantee on the downstream performance (Theorem \ref{thm:general-generalization-gap}). In the following part, we will present a new framework to analyze this variance term theoretically.

\subsection{A New Theoretical Analysis Framework under Augmentation Overlap}

As analyzed in Theorem \ref{thm:general-generalization-gap}, when contrastive learning has enough negative samples, the gap between pretraining and downstream performance mainly hinges on the variance of features in the same class. However, with current analysis tools, it is quite difficult to analyze the variance of intra-class samples. Therefore, in this section, we build a new augmentation overlap based framework to do this, and further theoretically characterize the generalization of contrastive learning.

Intuitively, contrastive learning is an instance-level task that can not bridge the samples in the same class. However, as shown in Figure \ref{fig:clusted-features}, the features of the same class will be clustered while the features of different classes will be separated, which indicate the intra-class variance gradually decreases along with the training process. Observing the training process of contrastive learning, we find that the different samples of the same class will generate quite similar views. For example, as shown in Figure \ref{fig:my_label}(b), appropriate data augmentations will make the views of different cars focus on similar tires. Then the training process of contrastive learning will close the feature distance of two cars as they share a similar view. Thus, the data augmentations play a key role in the understanding of contrastive learning, especially the overlapped views of augmentations. 

\begin{figure}
    \centering
    \subfigure[Contrastive learning with an augmentation graph satisfying intra-class connectivity.]{
    \includegraphics[width=.4\textwidth]{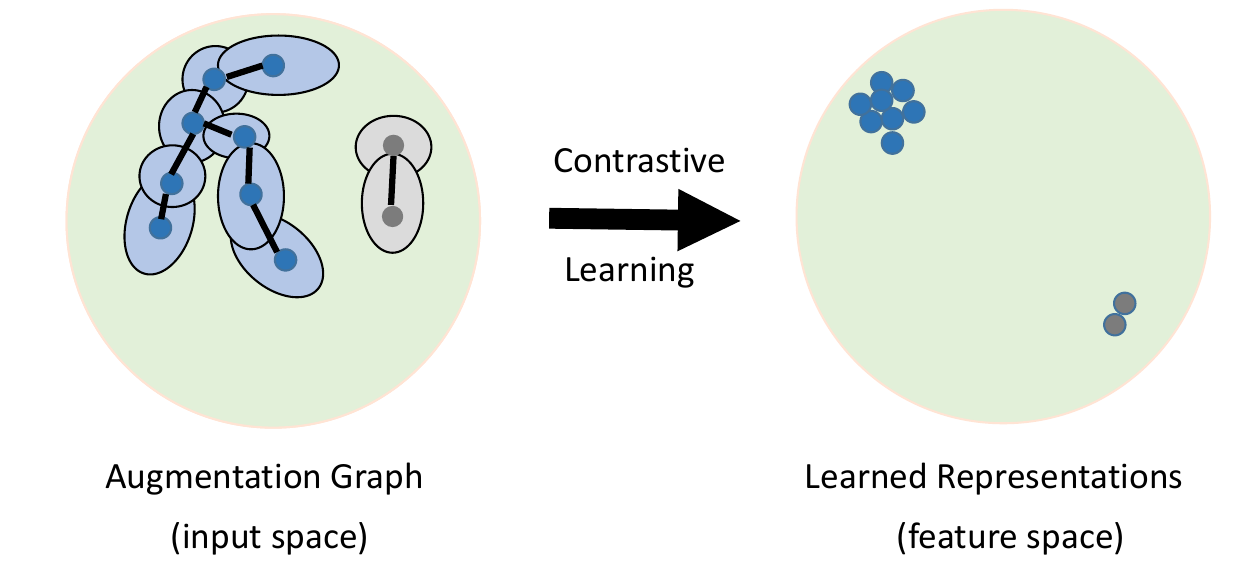}
    \label{fig:augmentation-graph}
    }
    \hfill
    \subfigure[Augmentation  graph  under  increasing  augmentation  strengthes (from left to right).]{
    \includegraphics[width=.55\textwidth]{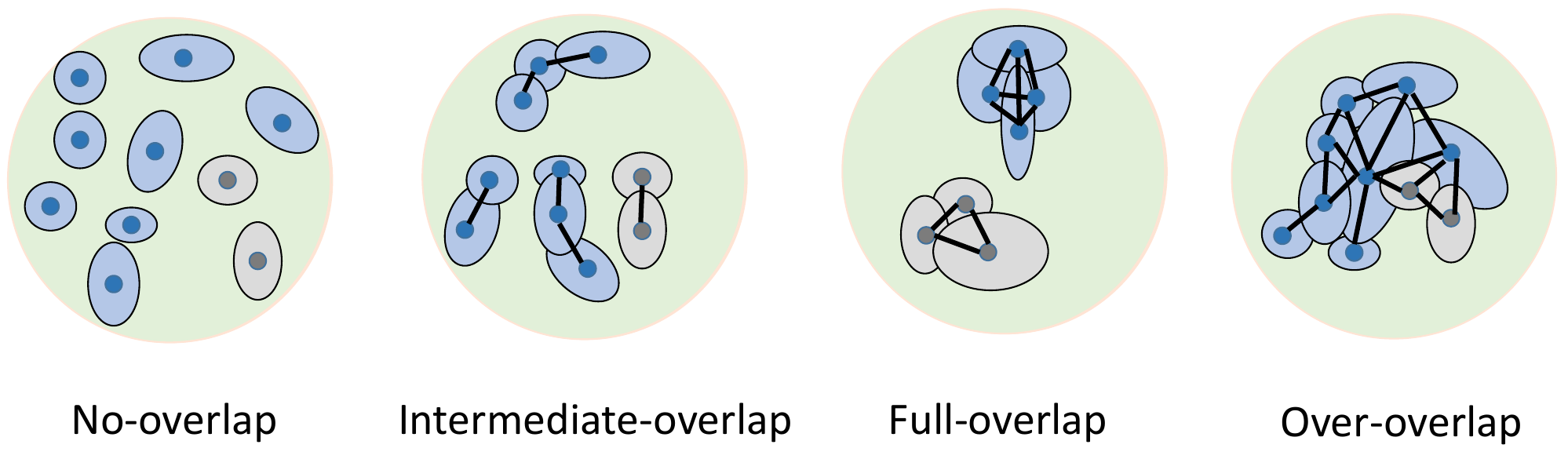}
    \label{fig:aug-strength}
    }
    \caption{Illustrative examples of augmentation graphs, where each dot denotes a sample $x\in\gD_u$ and its color denotes its class. The lighter disks denote the support of the positive samples $p(x^+|x)$. We draw a solid edge for each pair that has an edge. 
    }
    \label{fig:augstrength visulization}
\end{figure}

Accordingly, we formulate the above intuitive understanding of augmentation overlap mathematically via a graph. 
Assume a common graph $\gG(V,E)$ is composed of two components: a set $V$ representing the vertices and a set $E$ representing the edges between the vertices. Combined with a set of augmentations $\gT=\{t\mid t:\sR^d\to\sR^d\}$, an augmentation graph is defined as follows:
\begin{Definition}[Augmentation Graph] Given unlabeled { natural} data $\gD_u = \{\bar{x_i}\}_{i=1}^{N}$ and a collection of augmentations $\gT=\{t\mid t:\sR^d\to\sR^d\}$, an augmentation graph $\gG(V,E(\gT))$ is defined as 
\begin{itemize}
    \item its vertices are the samples, \ie $V=\{\bar{x_i}\}_{i=1}^N$;
    \item its edge $e_{ij}$ between two vertices $\bar{x_i}$ and $\bar{x_j}$ exists when they have overlapped views, \ie there exist two augmentations $t_1,t_2 \in \gT$ satisfying $t_1(\bar{x_i}) = t_2(\bar{x_j})$.
\end{itemize}
\end{Definition}
Based on the augmentation graph, we propose to utilize the properties of graph, \ie connectivity, to replace the impractical conditional independence assumption. Specifically, an augmentation graph is connected when its any two vertices (\eg $\bar{x_i},\bar{x_j}$) are connected, \ie there exists a path $(\bar{x_i},\cdots,\bar{x_j})$ between the two vertices. An illustrative example is shown in Figure \ref{fig:augmentation-graph}.

\begin{assumption}
Let $\train_k$ be the set of the samples in the class $k$ of $\train_u$. There exists an appropriate augmentation set $\gT$ satisfying that, for $k \in \{1,\cdots,K\}$, the augmentation graph $\gG(\train_k,E(\gT))$ is connected.
\label{ass:intra-class connectivity}
\end{assumption}
The assumption says that, for every pair of the intra-class samples, we only assume the path exists between them rather than requiring direct edges, \ie we only need connected graphs instead of complete graphs, which makes the connected augmentation graph assumption more practical. 

Beside the assumption on intra-class samples, we also need an assumption on the learned encoder of contrastive learning, that is, the distance between the positive samples will gradually converge to a small constant $\varepsilon$ during the contrastive learning process as demonstrated in \citet{wang}. 
\begin{assumption}
$\forall\ x,x^+ \sim p(x,x^+)$, we assume the learned mapping $f$ by contrastive learning is $\varepsilon$-alignment, \ie$\|f(x) - f(x^+)\| \leq \varepsilon$.
\label{def:epsilon-alignment}
\end{assumption}

With the assumptions on the intra-class connectivity of augmentation graph and the characterization on the alignment of positive pairs, we have the following theorem:
{
\begin{Theorem}[Guarantees with Connected Augmentation Graph] If Assumptions \ref{assumption:label-consistency} and \ref{ass:intra-class connectivity} hold, then $\forall f\in\gF$ satisfying $\varepsilon$-alignment (Assumption \ref{def:epsilon-alignment}), its classification risk can be upper and lower bounded by its contrastive risk as
\begin{equation}
\begin{aligned}
&\bar\gL_{\rm contr}(f)- (2+\frac{D\varepsilon}{2})D\varepsilon-4\sqrt{\alpha}-\frac{e}{\sqrt{M}}\\
& \qquad \leq\bar\gL_{\rm mCE}({f})
\leq\bar\gL_{\rm contr}(f) + 2D\varepsilon +4\sqrt{\alpha} + \frac{e}{\sqrt{M}},
\end{aligned}
\label{eq:weak-ge-gap}
\end{equation}
where $D$ denotes the maximal radius of the intra-class augmentation graphs $\{\gG_k,k=1,\dots,K\}$.
\label{thm:weak-generalization-gap}
\end{Theorem}
}

Comparing Eq. \ref{eq:general-gap} and Eq. \ref{eq:weak-ge-gap}, we can see that the intra-class variance term $\sqrt{\var(f(x)\mid y)}$ is replaced by $D\varepsilon$, where $D$ is the radius of augmentation graph and $\varepsilon$ measures the feature distance between positive samples. Intuitively, for samples in each class $k$, if its corresponding augmentation graph $\gG_k$ is connected, there exists a path between any two samples of the same class and aligning the adjacent sample pairs in the paths can eventually align all the samples of the same class. As a result, the intra-class variance can be controlled. 
Therefore, to ensure the generalization of contrastive learning, we need to meet two conditions: 1) choosing appropriate data augmentations to construct a connected augmentation graph with a smaller radius (smaller $D$); and 2) selecting a proper objective to align the positive samples better (smaller $\epsilon$). In Figure \ref{fig:augmentation-graph}, we visualize the training process of contrastive learning under augmentation graph.

{
\begin{Corollary}
If Assumptions \ref{assumption:label-consistency} and \ref{ass:intra-class connectivity} hold, then, for $f$ satisfying $0$-alignment, its classification risk can be upper and lower bounded by its contrastive risk as
\begin{equation}
\begin{aligned}
&\bar\gL_{\rm contr}(f )-4\sqrt{\alpha}-\frac{e}{\sqrt{M}}  \leq&\bar\gL_{\rm mCE}({f })\leq\bar \gL_{\rm contr}(f )+4\sqrt{\alpha}+ \frac{e}{\sqrt{M}}.
\end{aligned}
\end{equation}
\label{thm:closed-generalization-gap}
\end{Corollary}
}
The corollary shows that the intra-class variance term will varnish when the positive samples are perfectly aligned ($\epsilon = 0$). Furthermore, if enough negative samples are given ($M \to \infty$) { and we design appropriate augmentations that do not change the labels of samples ($\alpha=0$)}, the gap between pretraining and downstream performance will be totally closed, \ie the unsupervised contrastive learning is equivalent to a supervised task.

\subsection{Further Discussion on the Connectivity of Augmentation Graph}
Theorem \ref{thm:weak-generalization-gap} has provided a view of graph radius for the characterization of the connectivity of augmentation graph. In this part, we additionally provide another new perspective, \ie spectral property of augmentation graph.

{
\begin{Corollary}
If Assumptions \ref{assumption:label-consistency} and \ref{ass:intra-class connectivity} hold, then $\forall f\in\gF$ satisfying $\varepsilon$-alignment, its classification risk can be upper and lower bounded by its contrastive risk as
\begin{equation}
\begin{aligned}
&\bar\gL_{\rm contr}(f)- \left(2+\frac{\log((1-\omega^2)/\omega^2)}{2\log(\lvert \lambda_1 \rvert /\lvert \lambda_2 \rvert )}\varepsilon\right)\frac{\log((1-\omega^2)/\omega^2)}{\log(\lvert \lambda_1 \rvert /\lvert \lambda_2 \rvert )}\varepsilon-4\sqrt{\alpha}-\frac{e}{\sqrt{M}}\\
& \qquad \leq\bar\gL_{\rm mCE}({f})
\leq\bar\gL_{\rm contr}(f) + \frac{2\log((1-\omega^2)/\omega^2)}{\log(\lvert \lambda_1 \rvert /\lvert \lambda_2 \rvert )}\varepsilon +4\sqrt{\alpha}+ \frac{e}{\sqrt{M}},
\end{aligned}
\label{eq:spectral}
\end{equation}
where $\omega =  {\rm min}_{i,k}\lvert (\mu_{1k})_i \rvert,$ $\lambda_1 =  {\rm min}_k \vert\lambda_{1k}\vert$ and $\lambda_2 =  {\rm max}_k \vert\lambda_{2k}\vert$. Among them, $\mu_{1k},\mu_{2k},\cdots$ are the orthonormal eigenvectors with eigenvalues $\lambda_{1k},\lambda_{2k},\cdots$ $( \lvert \lambda_{1k} \rvert \geq \lvert \lambda_{2k} \rvert \geq \cdots)$ of the adjacent matrix of intra-class subgraph $\gG_k$.  
\label{cor:spectral}
\end{Corollary}
}
Comparing Eq. \ref{eq:weak-ge-gap} and Eq. \ref{eq:spectral}, we find that the graph radius $D$ is replaced by $\frac{\log((1-\omega^2)/\omega^2)}{\log(\lvert \lambda_1 \rvert /\lvert \lambda_2 \rvert )}$ that is controlled by the largest and second largest eigenvalues of the augmentation graph. The eigenvalues are widely discussed in the spectral graph theory and a smaller $|\lambda_2|$ implies the stronger connectivity of the graph \citep{10.2307/1990973}. From the corollary, we note that the gap between the pretraining and downstream performance is narrowed when $\lambda_2$ decreases, which further verifies that contrastive learning needs the augmentation graph to be closely connected. 

Previously, there is another work analyzing the downstream performance of contrastive learning from a spectral graph perspective \citep{haochen2021provable}, but the difference can be obviously observed: 1) our analysis is applicable for the widely adopted InfoNCE and CE losses, while theirs is developed for their own spectral loss; 2) ours starts from the alignment and uniformity perspective while theirs starts from the matrix decomposition perspective; and 3) there are several kinds of cases that their method fails to analyze while ours can. As the former two points are easily verified, we only provide the detailed comparison on the third point below.

{
\begin{Lemma}[The main results in \citet{haochen2021provable}]
Let  $f^\star\in\gF:\sR^d\to\sS^{m-1}$ be a minimizer of $\gL_{\rm{contr}}(f)$ and $m \geq 2K$. The downstream error is denoted as $Err(f^\star) = \E_{p(x,y)}(\hat y(x)\neq y)$ where $\hat y(x) =\argmax g(f^\star(x))$ is the predicted label with the downstream classifier $g$. Then, we have
\begin{equation}
    Err(f^\star) \leq \gO\left(\frac{\alpha}{\rho^2_{\lfloor m/2 \rfloor}}\right),
\label{eq:haochen}
\end{equation}
where $m$ is the representation dimension, $K$ is the number of classes, $\alpha$ is the minimum error under a set of labeling functions where the augmented views and the anchor sample have different labels, and $\rho_q$ is the sparsest $q$-partition of augmentation graph.
\label{thm:spectral bound}
\end{Lemma}
}
Their bounds (Lemma \ref{thm:spectral bound}) and ours (Corollary \ref{cor:spectral}) are both based on the connectivity of augmentation graph but with different measures. We use the first and the second largest eigenvalues of the adjacent matrix of the augmentation graph while they use the sparsest $q$-partition that is estimated by the $q$-smallest eigenvalues of the normalized Laplacian matrix of the augmentation graph. Thus, these two bounds are not identical: 1) ours provide both the upper and lower bounds while they only provide the upper bound; and 2) there are some cases their method fails to analyze while ours could as shown below.

\textit{Case I:} If Assumption \ref{ass:intra-class connectivity} holds and $\lfloor m/2 \rfloor \leq K$, then $\rho_{\lfloor m/2 \rfloor} = 0$. When Assumption \ref{assumption:label-consistency} holds, then $\alpha = 0$. The bound in Lemma \ref{thm:spectral bound} becomes a $\frac{0}{0}$ term which fails to be analyzed.

\textit{Case II:} If we adopt the same setting in Proposition \ref{prop:wang-couterexample}, \ie there exists no augmentation overlap for inter-anchor samples, the bound in Lemma \ref{thm:spectral bound} becomes a $\frac{0}{0}$ term again.
\\~\\
While for our bounds, the above two cases can still be characterized (proofs are shown in Appendix). This is because that our bounds are not dependent on the output dimension $m$ and can still analyze the difference between representations when $\alpha = 0$. That is, we can analyze the downstream performance of contrastive learning in a more fine-grained way with less restrictions.

\section{Analysis of Augmentation Strategy on Augmentation Overlap}
\label{sec:effect of augmentation graph}
Based on the above analysis, we can find that the gap between pretraining and downstream tasks hinges on the connectivity of the augmentation graph $\gG(V, E(T))$. 
In this section, we will further analyze how the data augmentation strategy in contrastive learning influences the graph connectivity (augmentation overlap).

\subsection{Degree of Augmentation Overlap}
\label{subsec:thm on random graph}
To simplify our analysis, here we consider the following setting. For each class $k$, there is a cluster center $c_k$ on a hypersphere $\sS^d$ where $N$ anchor samples are uniformly distributed around $c_k$. The positive samples are obtained by adding a uniform noise sampling from a uniform distribution $U(0,r)^d$ to the anchor samples. Intuitively, when the augmentation strength $r$ is too weak, there exist no overlapped views across different samples. When the augmentation is too strong, samples of different classes may be aligned together. Formally, based on the random graph theory, we have the following theoretical results. 

\begin{Theorem}
For $N$ random samples taken from a class, when gradually increasing the augmentation strength $r$, we have the following degree of augmentation overlap:
\begin{enumerate}[label=(\alph*)]
    \item \textbf{No-overlap.} When $0\leq r < r_1=\frac{[(d/2)!]^{\frac{1}{d}}}{\sqrt{\pi}}{(\frac{1}{d})!}(\frac{S}{N-1})^{\frac{1}{d}}[1-\frac{1/d+1/d^2}{2(N-1)}+O(\frac{1}{(N-1)^2})]$ where $S$ is the surface area of sample distribution and $r_1$ is the minimal pairwise distance among $N$ samples, all samples (vertices) in the augmentation graph are isolated. As a result, the learned features could be totally random as in Proposition \ref{prop:wang-couterexample}. 
    \item \textbf{Over-overlap.} When $r\geq r_3=\frac{1}{2}\min_{i,j}\Vert c_i-c_j\Vert$ where $r_3$ is the (asymptotic) minimal distance between samples from different classes, the label consistency assumption is no longer guaranteed. 
    \item \textbf{Full-overlap.} When
    $r_2 \le r < r_3$ where 
    $r_2=\frac{[(d/2)!]^{\frac{1}{d}}}{\sqrt{\pi}}\frac{(N-2+1/d)!}{(N-2)!}(\frac{S}{N-1})^{\frac{1}{d}}[1-\frac{1/d+1/d^2}{2(N-1)}+O(\frac{1}{(N-1)^2})]$ is the maximal pairwise distance among $N$ samples, all samples from the same class are directly connected while samples from the different classes are not connected. As a result, the augmentation graph for each class is a complete graph. 
    \item \textbf{Intermediate-overlap.} When $r_1\leq r < r_2$, there exists at least two samples that are connected, while the whole augmentation graph is not a complete graph. 
\end{enumerate}
Under cases (c) and (d), the classwise connectivity in Assumption \ref{ass:intra-class connectivity} is guaranteed. 
\label{theorem:augmentation-strength}
\end{Theorem}

The above Theorem \ref{theorem:augmentation-strength} indicates a trade-off on the augmentation strength. On the one hand, we need the augmentation to be strong enough to align the samples from the same class. On the other hand, we need to avoid too strong augmentations that generate overlapped views for inter-class samples. In Figure \ref{fig:aug-strength}, we visualize the relationship between different degrees of overlap and the connectivity of augmentation graph.
According to Theorem \ref{thm:weak-generalization-gap}, the guarantees about the generalization of contrastive learning only need the augmentation graph to be connected (may not need to be a complete graph). Thus, there is a proper (perfect) augmentation strength from `intermediate-overlap' to `full-overlap', named `perfect-overlap', which can be described through the following theorem.

\begin{Theorem}
Denote the minimal augmentation strength needed for connectivity as: $r_{mc}=\inf \{r_1 \leq r \leq r_2:\gG(V,E(\gT)) \text{ is connected}\}$
and the volume of unit hyperball as $V_u$, we obtain:
\begin{equation}
\text{For } d \geq 2, r_{mc} = O\left((2\frac{(1-1/d)S\log N}{V_u N^2})^{\frac{1}{d}}\right) \text{as } N \to \infty.
\end{equation}
\label{theorem:augmetantion-infinity}
\end{Theorem}
Theorem \ref{theorem:augmetantion-infinity} implies that the perfect augmentation strength (making augmentation graph be connected) is strongly related to the sample number ($N$) and sample dimension ($d$). As the sample dimension increases, the needed augmentation strength to meet `perfect-overlap' will increase in an exponential order. Since the dimension of the samples in real-world data sets is quite large, stronger data augmentations are usually needed. However, simple data augmentations like Gaussian noise with strong variance (strong data augmentations) will change the label of images, which contradicts the label consistency assumption. Thus, we need to design data augmentations that are strong enough while do not change the label of samples simultaneously. This is in accordance with the practical fact that common contrastive learning methods usually use complex data augmentations like RandomResizedCrop and ColorJitter instead of simple Gaussian noise.

\subsection{Empirical Understanding of Augmentation Strength}
Based on the setting in Section \ref{subsec:thm on  random graph}, we conduct a series of experiments on a synthetic data set to verify the influence of the augmentation strength. Consider a binary classification task ($k=2$) with the InfoNCE loss. Data are generated from two uniform distributions on a unit ball $\sS^2$ in the $3$-dimensional space (one center is $(0,0,1)$ and another is $(0,0,-1)$). {More details can be found in Appendix \ref{sec: simulation rag}.} %

\begin{figure}[!t]
    \centering
    \includegraphics[width=\textwidth]{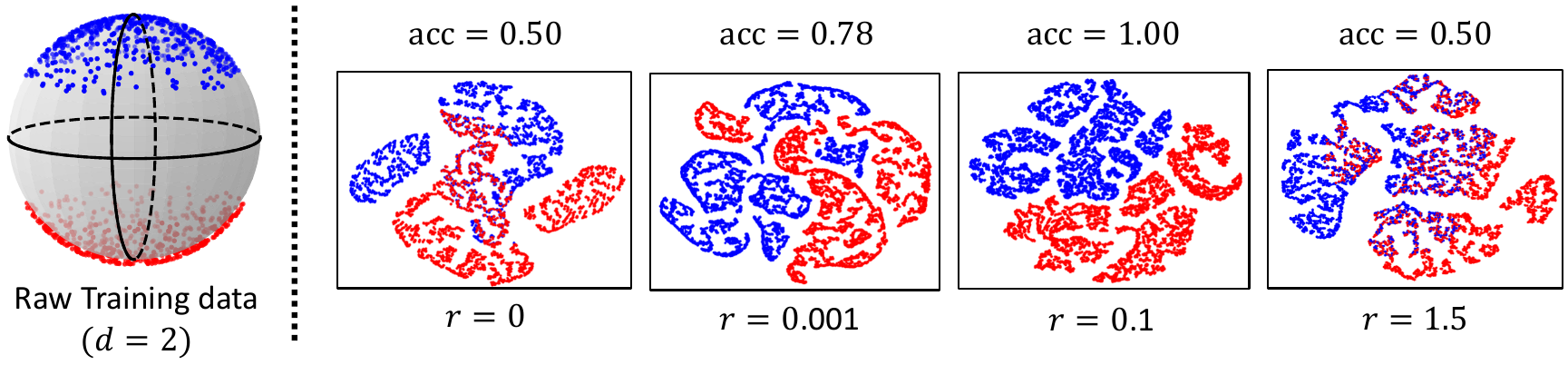}
    \caption{t-SNE visualization of features learned with different augmentation strength $r$ on the random augmentation graph experiment. Each dot denotes a sample and its color denotes its class.}
    \label{fig:random-tsne}
\end{figure}

As shown in Figure \ref{fig:random-tsne}, we find that with the enhancement of the augmentation strength $r$, the accuracy of the classification tasks increases first and then falls to 50\%. From the learned representations with different augmentation strengths, we find that when the augmentation strength is weak, the features of the intra-class samples are more isolated. When the augmentation strength is too strong, the features of the different classes are mixed. In both cases, it is almost impossible to obtain a linear classifier with good downstream performance.

\begin{figure}
    \centering
    \subfigure[$r=0.0$, \text{acc} = 0.50, $D = \inf$. ]{
    \includegraphics[width=0.17\textwidth]{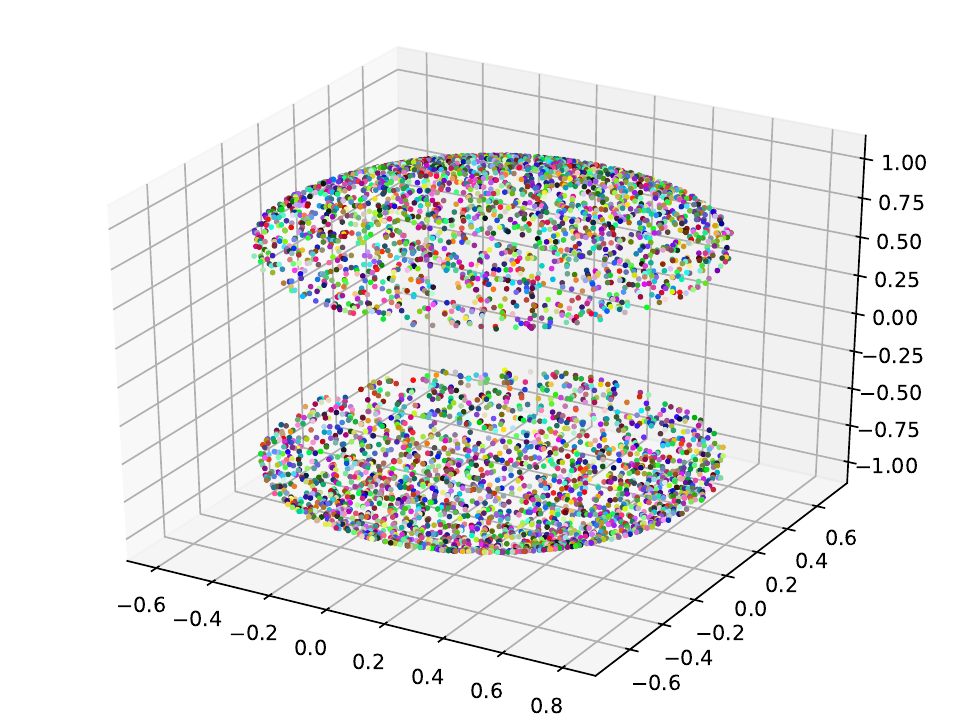}
    \label{fig:vis_syn_0}}
    \hfill
    \subfigure[$r=e^{-3}$, \text{acc} = 0.78, $D = \inf$. ]{
    \includegraphics[width=0.17\textwidth]{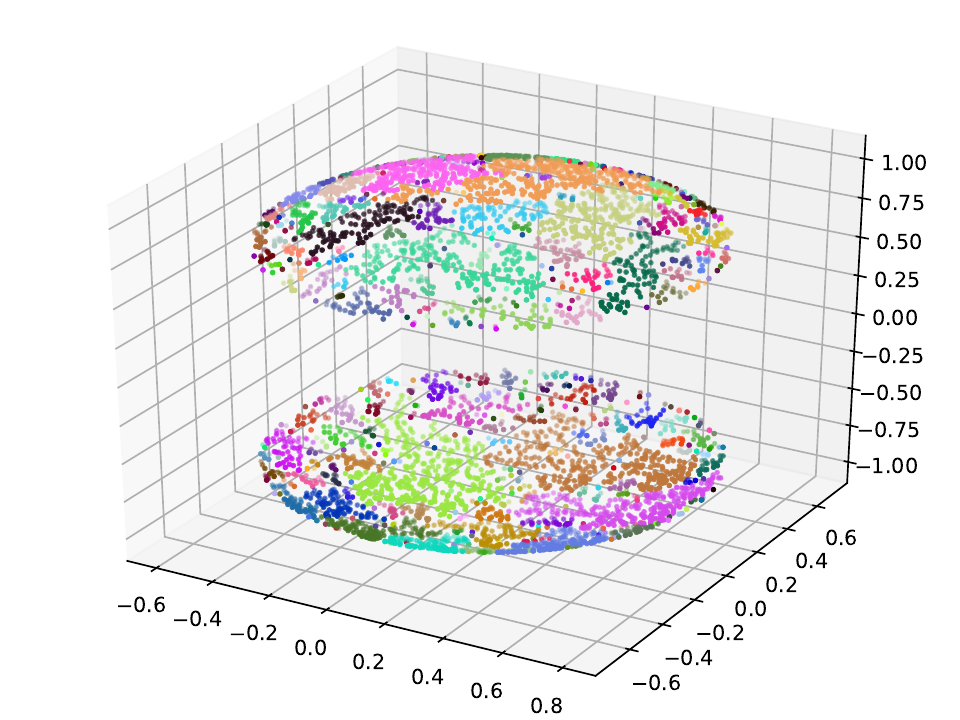}
        \label{fig:vis_syn_0_001}}
    \hfill
    \subfigure[$r=0.08$, \text{acc} = 0.96, $D=5$.]{
    \includegraphics[width=0.17\textwidth]{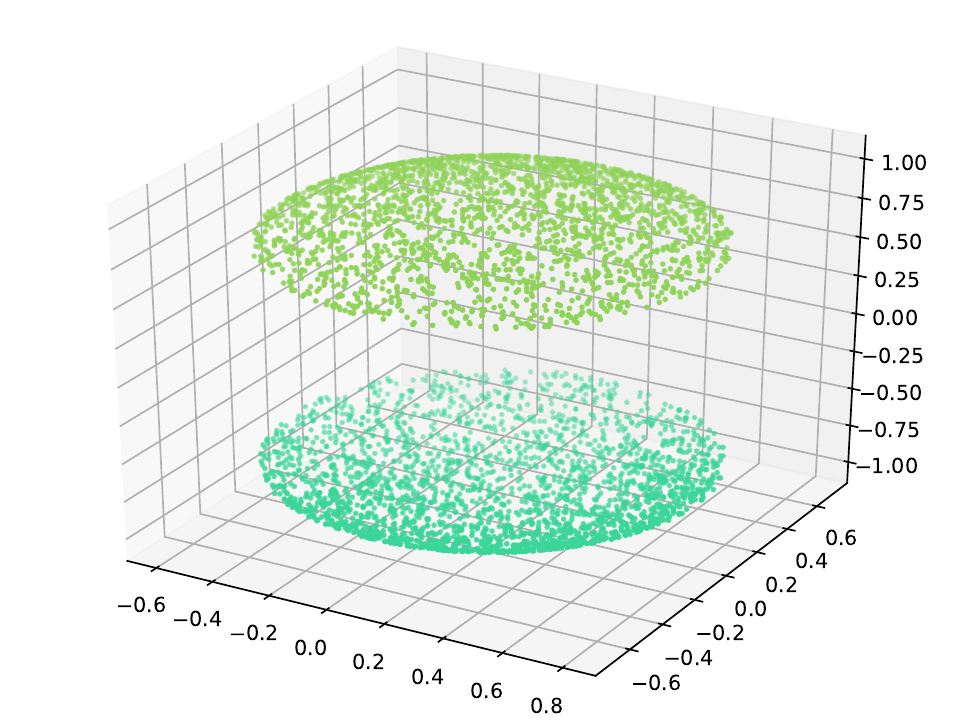}
        \label{fig:vis_syn_0.1}}
    \hfill
    \subfigure[$r=0.5$, \text{acc} = 1.00, $D=2$.]{
    \includegraphics[width=0.17\textwidth]{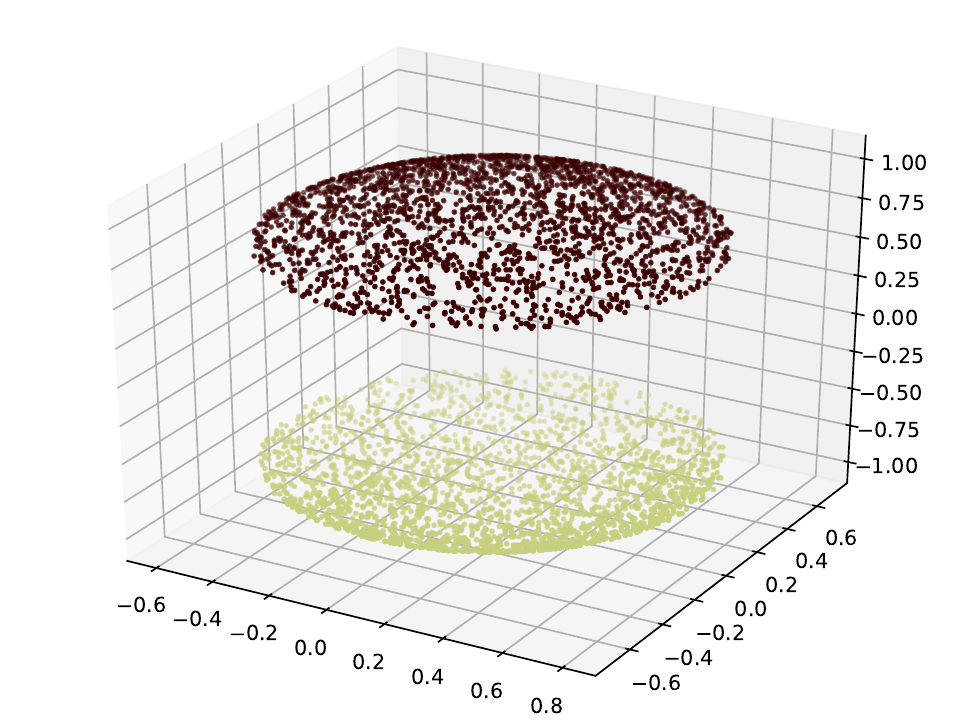}
        \label{fig:vis_syn_0.5}}
    \hfill
    \subfigure[$r=1.5$, \text{acc} = 0.50, $D=1$]{
    \includegraphics[width=0.17\textwidth]{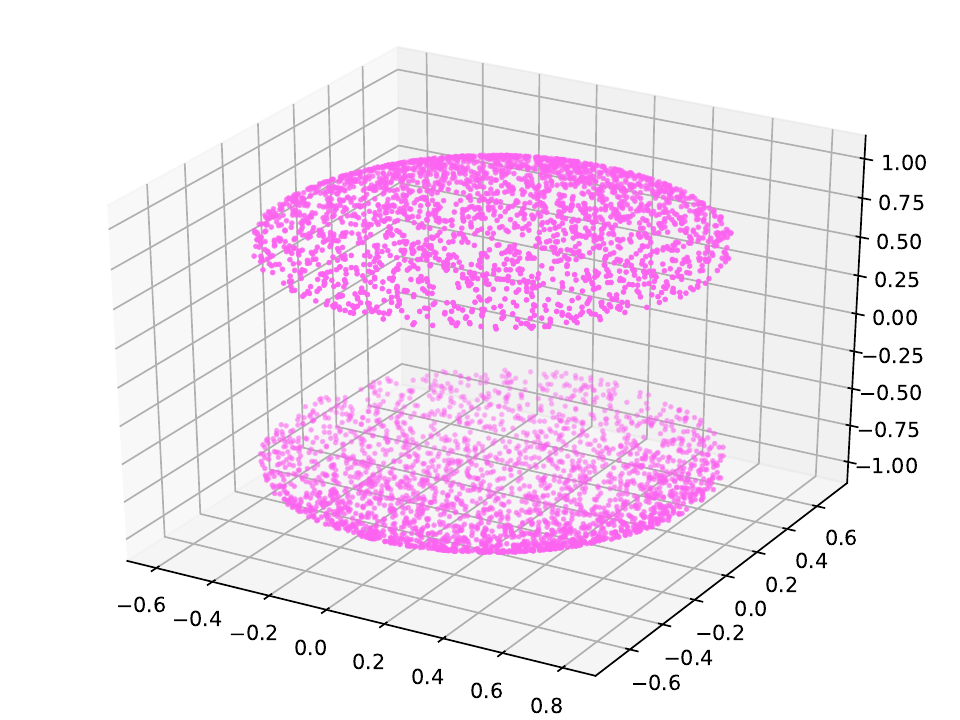}
        \label{fig:vis_syn_1.5}}
    \caption{Visualization of the augmentation graph with different augmentation strength $r$ on the synthetic data. Each color denotes a connected component. The number of the connected components and the maximal intra-class radius $D$ decrease with the increase of the augmentation strength.}
    \label{fig:synthetic-augmentation-graph}
\end{figure}

To further understand the relationship between the strength of augmentations and the connectivity of augmentation graph, we visualize the augmentation graphs with different augmentation strengths on the above synthetic data set and present the maximal intra-class radius $D$ of them in Figure \ref{fig:synthetic-augmentation-graph}. We observe that when augmentations are not applied on the data (Figure \ref{fig:vis_syn_0}), all the samples are isolated ($D=\inf$) and contrastive learning becomes a simple instance-level discriminative task. The degree of intra-class overlap aligns with the increase of the augmentation strength. At the point of the highest linear accuracy where $r=0.5$ (Figure \ref{fig:vis_syn_0.5}), all the intra-class samples are closely connected ($D=2$) while all the inter-class samples are separated, which satisfies Assumption \ref{ass:intra-class connectivity}. When the augmentation is too strong (Figure \ref{fig:vis_syn_1.5}), we find that the two subgraphs of the augmentation graph are connected, which means the inter-class samples may be wrongly aligned, and as a result, the linear accuracy dramatically decreases to 50\%.

\begin{figure}
    \centering
    \subfigure[Almost no-overlap augmentation graph (r=(0.99,1), {acc}=0.25).]{
    \includegraphics[width=0.3\textwidth]{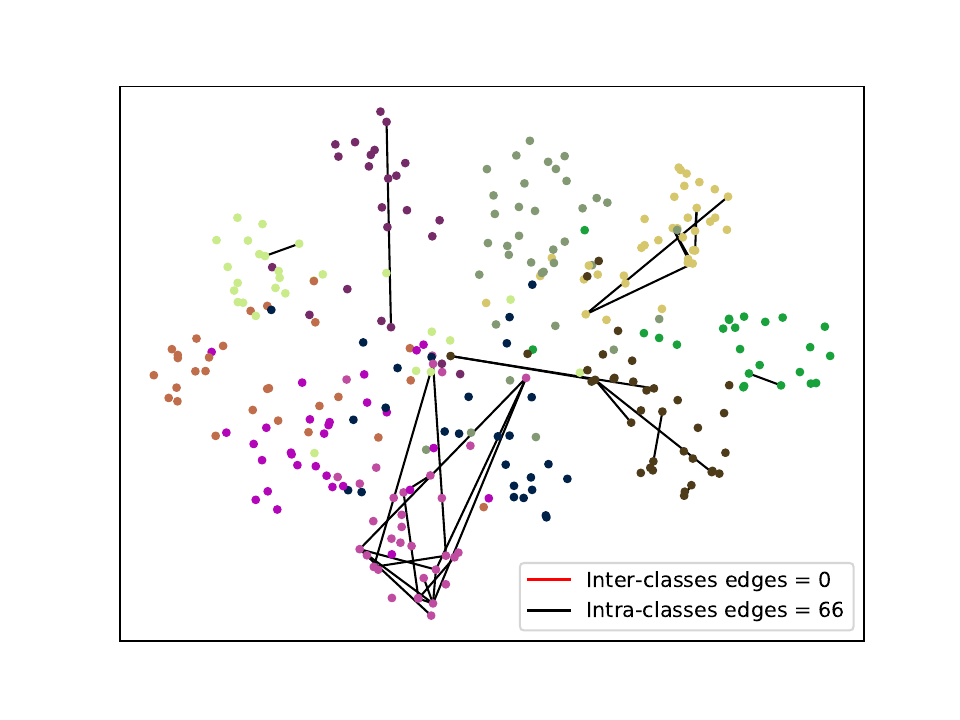}
    }
    \hfill
    \subfigure[Proper-overlap augmentation graph (r=(0.08,1), {acc}=0.75).]{
    \hfill
    \includegraphics[width=0.3\textwidth]{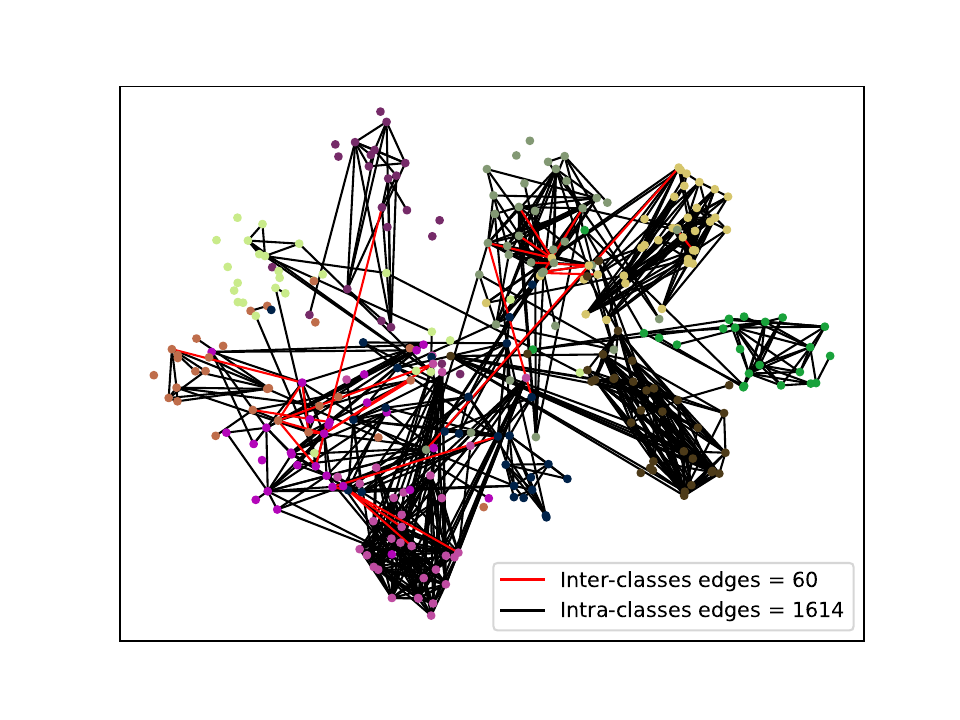}}
    \subfigure[Over-overlap augmentation graph (r=(0.01,0.03), {acc}=0.29).]{
    \includegraphics[width=0.3\textwidth]{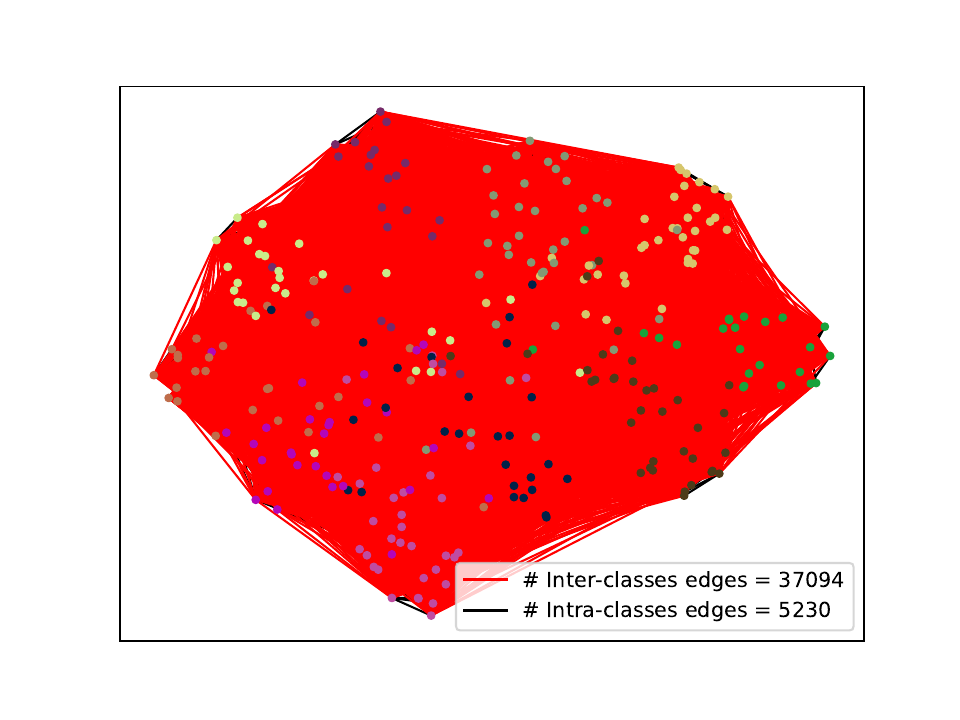}}
    \caption{The augmentation graph of CIFAR-10 with different strength $r$ of RandomResizedCrop. We choose a random subset of test images and randomly augment each one 20 times. Then, we calculate the sample distance in the representation space as in prior work like FID \citep{FID} and draw edges for image pairs whose smallest view distance is below a small threshold. Afterwards, we visualize the samples with t-SNE and color intra-class edges in \textbf{black} and inter-class edges in \textbf{\textcolor{blue}{blue}} and report their frequencies.
    }
    \label{fig:cifar-10-augmentation-graph}
\end{figure}
 
Besides the synthetic data set, we also verify the practicality of our theoretical results on the real-world data set CIFAR-10. {{While the augmentation graph is defined in terms of connections in the input space, directly measuring semantic similarity in this high-dimensional space presents significant challenges. To overcome this, we adopt a well-established surrogate metric by using pretrained representations to approximate the similarity of input images, following \citet{dwibedi2021little}.}}
Specifically, for characterizing the connectivity of two samples, we augment each sample 20 times and then compute their cosine similarity. An edge is considered to exist between two samples if the cosine similarity between their augmented counterparts exceeds a threshold. We use the RandomResizedCrop as the data augmentation and regard the scale of the crop $(a,b)$ as the augmentation strength $r$, \ie $r = (1-a) + (1-b)$. The connectivity of the augmentation graph on CIFAR-10 is visualized in Figure \ref{fig:cifar-10-augmentation-graph}, we find that when the augmentation strength is too weak where the scale of the crop is (0.99,1) (Figure \ref{fig:cifar-10-augmentation-graph}(a)), the linear accuracy is 25\% and most samples are isolated. When we adopt appropriate augmentation strength (Figure \ref{fig:cifar-10-augmentation-graph}(b)), the linear accuracy increases to 75\% and at the same time, the edges between intra-class samples take 96.4\% of all the edges. When the augmentation strength is too strong (Figure \ref{fig:cifar-10-augmentation-graph}(c)), the linear accuracy falls down to 29\% and the edges of the inter-class samples take 87.6\% of the edges. The empirical findings in the real-world data sets are quite close to the results on the synthetic data sets. The above results further verify that the connectivity of augmentation graph decides the learned representation of contrastive learning, \ie too weak or too strong data augmentations will hurt its downstream performance.

\section{Applications of Augmentation Overlap Theory}
\label{sec:ARC total}

Although contrastive learning has received impressive empirical success in many downstream tasks, how to quickly evaluate the quality of the learned representation is still under-explored. In practice, the most common method is to train a linear classifier following the pretrained encoder, \ie linear evaluation \citep{simclr,moco,BYOL}. However, linear evaluation needs supervised label information and extra training time which are quite expensive for large-scale real-world data sets. Recently, some works try to evaluate the performance of contrastive learning without supervised labels, for example, \citet{selfaugment} find rotation prediction accuracy of the learned representation is strongly related to the classification accuracy in downstream tasks. However, their method still needs additional training time for the rotation classifier. In this section, we propose an unsupervised metric based on our augmentation overlap theory which can evaluate the performance of contrastive learning with almost no additional computational cost.

\begin{figure}[t]
    \centering
    \subfigure[$d_{out} > d_{in}$]{
    \includegraphics[width=.35\textwidth]{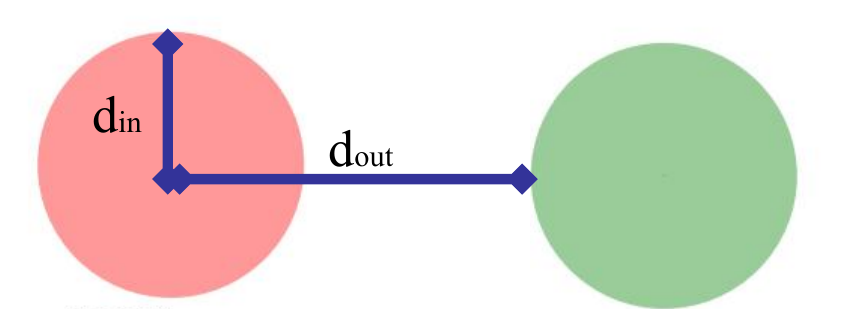}}
    \subfigure[$d_{out} = d_{in}$]{
    \includegraphics[width=.25\textwidth]{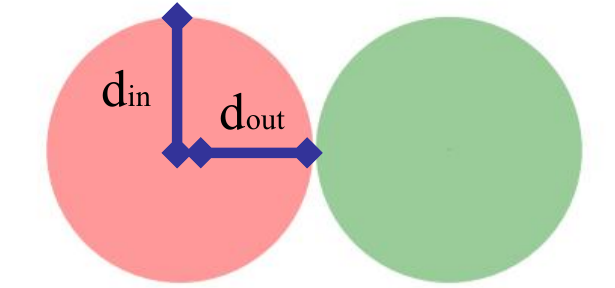}}
    \subfigure[$d_{out} < d_{in}$]{
    \includegraphics[width=.16\textwidth]{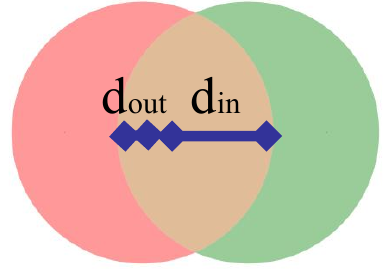}}
    \caption{{Measuring support overlap with the help of the distance between different augmented samples.}}
    \label{fig:ACR_intuition}
\end{figure}

\subsection{An Unsupervised Metric for Representation Evaluation}
\label{sec:ARC}
Based on our augmentation overlap theory, generating overlapped views of intra-class samples is a critical step of contrastive learning and the connectivity of the augmentation graph is strongly related to the downstream performance. Intuitively, when justifying whether two samples are adjacent in the augmentation graph, we need to enumerate all their augmented views to find whether they share an augmented view. However, going through the augmentation space needs huge computational costs which can hardly be implemented in practice, especially for the real-world data sets. To be specific, common contrastive learning paradigms adopt different kinds of data augmentations, including RandomResizedCrop, ColorJitter, GrayScale, and GaussianBlur. Each data augmentation has different continuous parameters, for example, Colorjitter controls the brightness, contrast, saturation, and hue of the images, which all can be random positive values. As the result, it is impractical to search the space of data augmentations with current computing resources.

\begin{figure}[t]
    \centering
    \subfigure[ACR \vs aug-strength r.]{
    \label{fig:ACR-augment-strength}
    \includegraphics[width=0.3\textwidth]{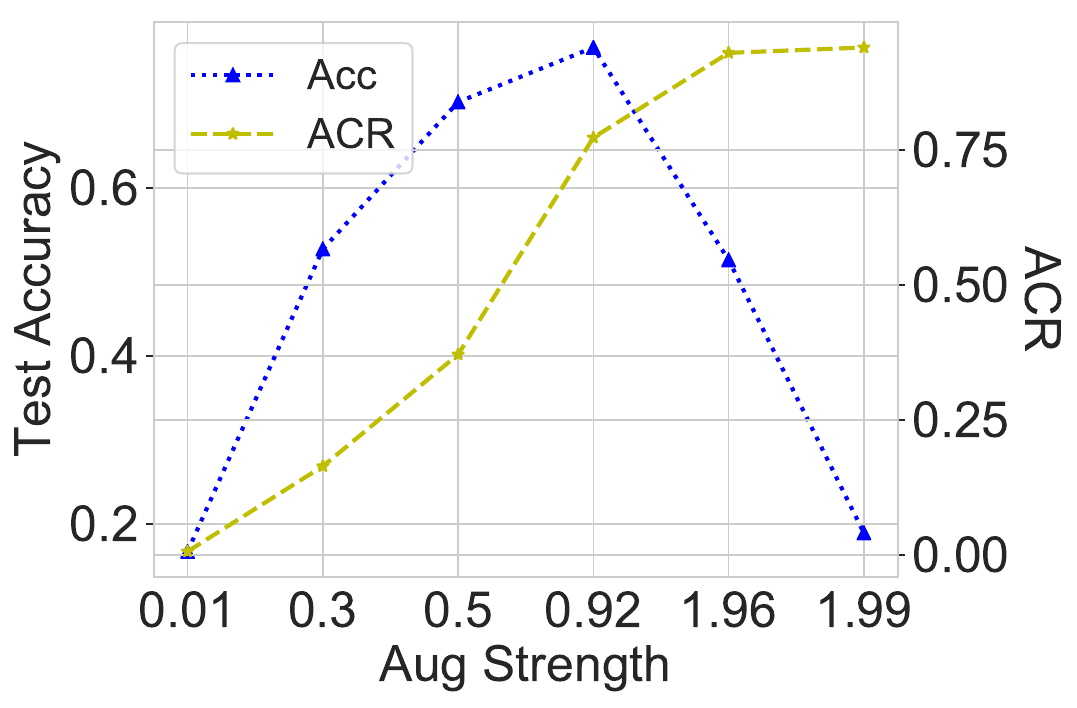}}
    \subfigure[ACR while training (r=0.01).]{
    \label{fig:ACR-r-0-01}
    \includegraphics[width=0.3\textwidth]{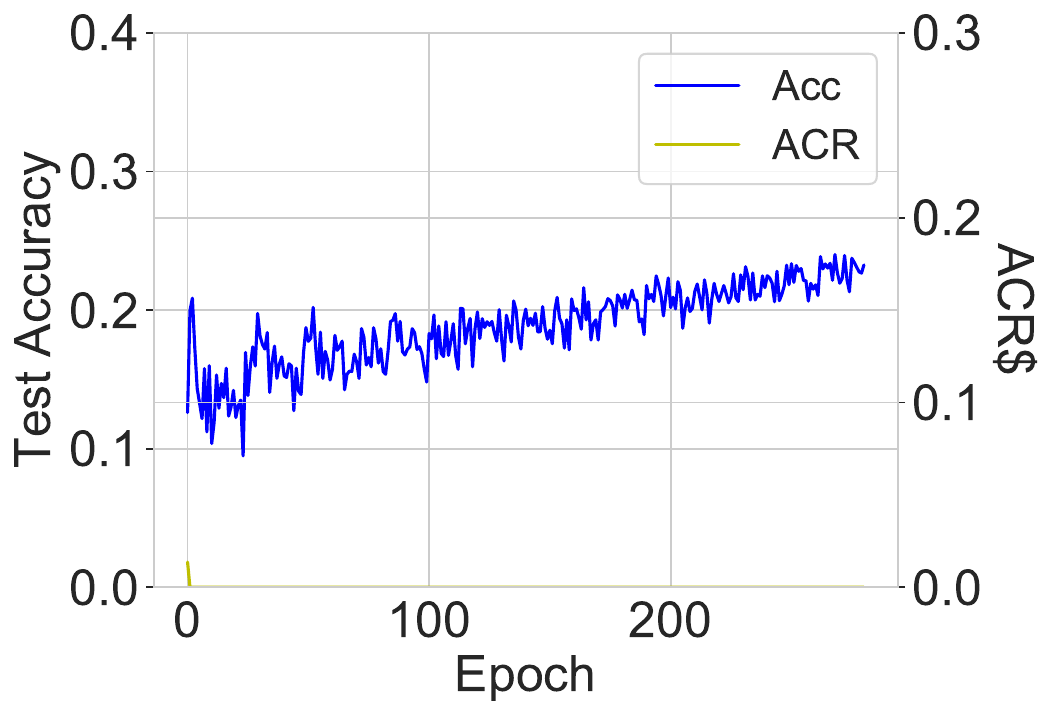}}
    \subfigure[ACR while training (r=0.92).]{
    \label{fig:ACR-r-0-92}
    \includegraphics[width=0.3\textwidth]{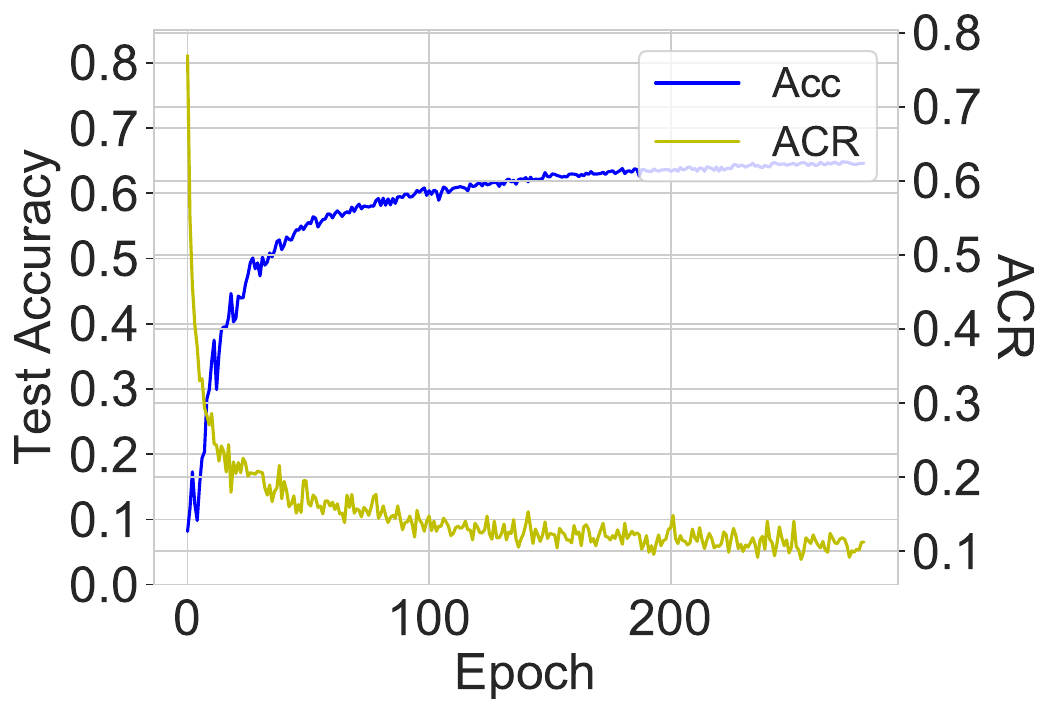}}
    \caption{(a) Average Confusion Rate (ACR) and downstream accuracy \vs different augmentation strength (before training). (b,c): ACR and downstream accuracy while training. }
    \label{fig:ACR}
\end{figure}

Therefore, we propose an approximation metric to measure the connectivity of augmentation graph with the help of the nearest neighbours around the sample. For each sample $x_i \in \gD_u$, we random augment it for $C$ times and get a support set $\hat{\gD}_u = \{x_{ij}, 0\leq i\leq N, 0\leq j \leq C\}$. We define $d_{in}(x_{ij},f) = \max\{\|f(x_{ij})-f(x_{ip})\|, 0\leq p \leq C\}$ as the maximal distance of the augmented  samples of the same anchor and $d_{out}(x_{ij},f) = \min\{\|f(x_{ij})-f(x_{lp})\|, 0\leq i,j \leq N, 0\leq p \leq C, i\neq l\}$ as the minimal distance of the augmented samples of different anchors in the feature space. As shown in Figure \ref{fig:ACR_intuition}, we find that when the different samples have support overlap, $d_{in}$ is larger than $d_{out}$. Therefore we can measure the degree of augmentation overlap by comparing $d_{in}$ and $d_{out}$. Following that, 
we define a Average Confusion Ratio (ACR) as the ratio of the intra-anchor distance is larger then the inter-anchor distance:
\begin{equation}
    \operatorname{ACR}(f)= \E_{x_{ij} \in \hat{\gD}_u}\mathbb{I}(d_{out}(x_{ij},f)\leq d_{in}(x_{ij},f)).
\end{equation}
Note that when $C\rightarrow \infty$ and the support set includes all anchors, ACR is exactly the ratio of the anchors that have overlapped views with other anchors. Intuitively, ACR approximately measures the ratio that augmented views of different anchors can be closer than the augmented views of the same sample, which means that different samples generate the overlapped views. To show the performance of ACR on real-world data sets, we first take the RandomResizedCrop operator with different strengths on CIFAR-10 as an example. The augmentation strength $r$ of RandomResizedCrop is defined as the scale range $[a,b]$ of crop, \ie $r = (1-a) + (1-b)$. As shown in Figure \ref{fig:ACR}(a), we find that ACR increases with stronger augmentations. The empirical results verify that the metric ACR is a closed approximator of the degree of the connectivity of the augmentation graph. However, Figure \ref{fig:ACR}(a) also shows that the downstream accuracy increases first and then falls down while ACR keeps increasing with stronger data augmentations, which is consistent with our analysis that when augmentation strength is too strong, there occurs over-overlap. As a result, only measuring ACR can not be a satisfying evaluation method. Figure \ref{fig:ACR}(b) and Figure \ref{fig:ACR}(c) show the change of ACR during the training process of contrastive learning. We find that when the augmentation strength is too weak, contrastive learning becomes a quite simple task, the initial ACR is low and it decreases to 0 in a short period, and when we adopt appropriate augmentation strength, the initial ACR is high and it will decrease mildly. Inspired by the empirical findings, we think the change process of ACR may be more strongly related to the downstream performance of contrastive learning, so we propose a new metric to estimate the downstream accuracy of contrastive learning without label messages, named ARC (Average Relative Confusion):
\begin{equation}
\operatorname{ARC}=\frac{1-\operatorname{ACR}(f_{\rm final})}{1-\operatorname{ACR}(f_{\rm init})}.
\end{equation}

To evaluate the effectiveness of ARC, we conduct comprehensive experiments on CIFAR-10 to demonstrate the relationship between ARC and downstream performance on the representations trained with different data augmentation and different strengths. {The details can be found in Appendix \ref{exp: ARc}.} 

\begin{figure}[t]
    \centering
        \subfigure[Different augmentations]{
    \includegraphics[width=0.3\textwidth]{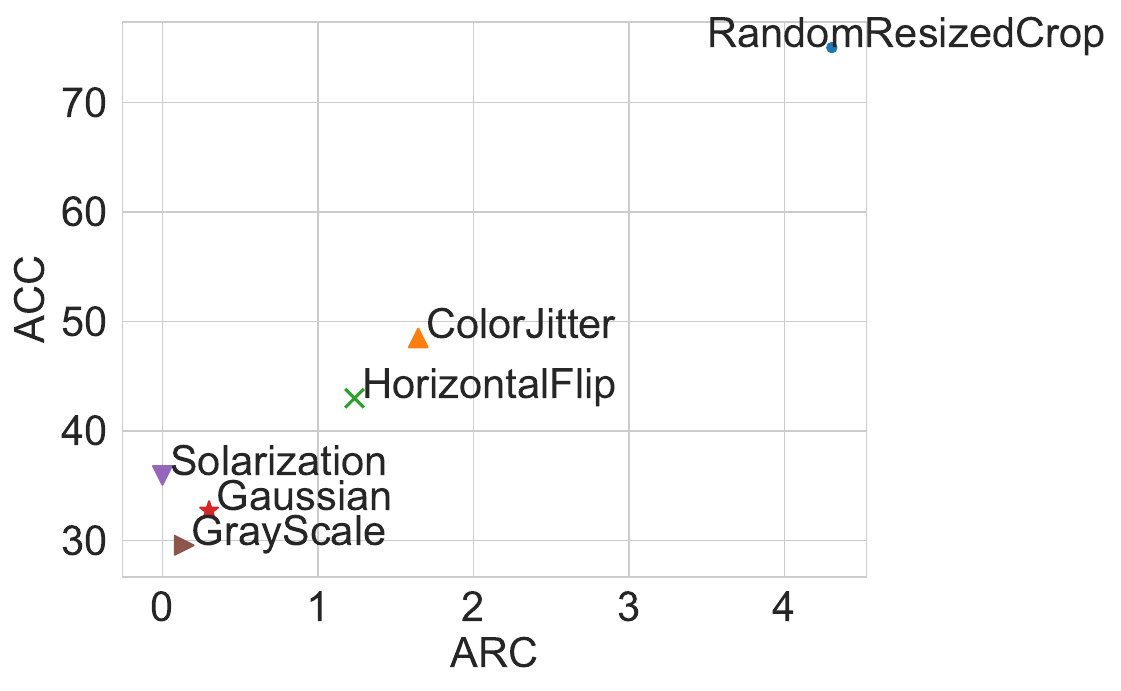}}
    \subfigure[RandomResizedCrop]{
     \includegraphics[width=0.3\textwidth]{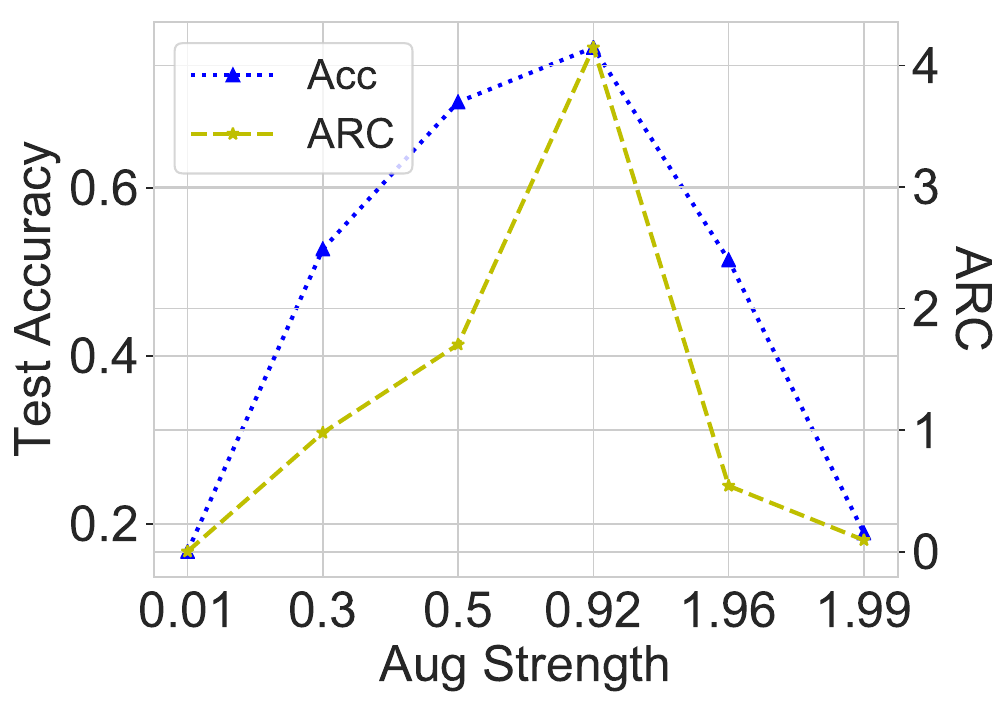}}
    \subfigure[Colorjitter]{
    \includegraphics[width=0.3\textwidth]{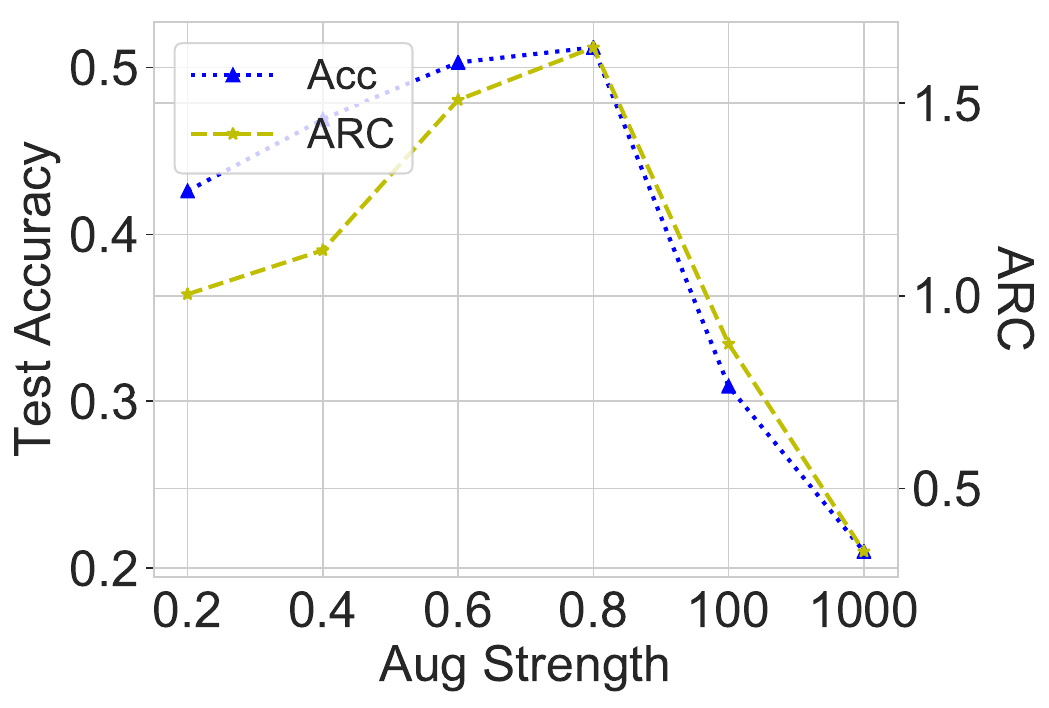}}

    \caption{Average Relative Confusion (ARC) and downstream accuracy \vs different data augmentations adopted in SimCLR with different strengths.}
    \label{fig:ARC}
    \vspace{-0.1in}
\end{figure}

\textbf{\textit{Different kinds of data augmentations.}} We test 6 kinds of augmentations used in SimCLR \citep{simclr}, \ie RandomResizedCrop, ColorJitter, GrayScale, HorizontalFlip, GaussianBlur and Solarization. Each augmentation is singly applied with the default parameter in SimCLR. The results are presented in Figure \ref{fig:ARC}(a), which shows that ARC aligns well with the linear accuracy of models trained by different kinds of augmentations. We also find that RandomResizedCrop and ColorJitter are the most two powerful data augmentations used in SimCLR, so we then focus on these two augmentations for further analysis.

\textbf{\textit{RandomResizedCrop with different strengths.}}
The strength $r$ of RandomResizedCrop is still defined as the scale $[a,b]$ of crop, \ie $r = (1-a) + (1-b)$. From Figure \ref{fig:ARC}(b), we find that the ARC curve and the linear accuracy curve have almost the same pace, \ie increasing first and then falling down, which indicates that our ARC metric has a close relationship with downstream performance.

\textbf{\textit{ColorJitter with different strengths.}} 
The strength of ColorJitter is directly controlled by four parameters, \ie brightness, contrast, saturation, and hue. Brightness, contrast, and saturation can be any positive values while hue is a positive value that is not larger than 0.5. So we set the parameter of ColorJitter as $(brightness, contrast, saturation, hue) = (r, r, r, \min(0.5,0.25*r) ) $. From Figure \ref{fig:ARC}(c), the curves of ARC and linear accuracy perform at almost the same pace again, which implies that our ARC metric is a good measure for the downstream performance evaluation. 

\begin{figure}[t]
    \centering
        \subfigure[Different augmentations]{
    \includegraphics[width=0.36\textwidth]{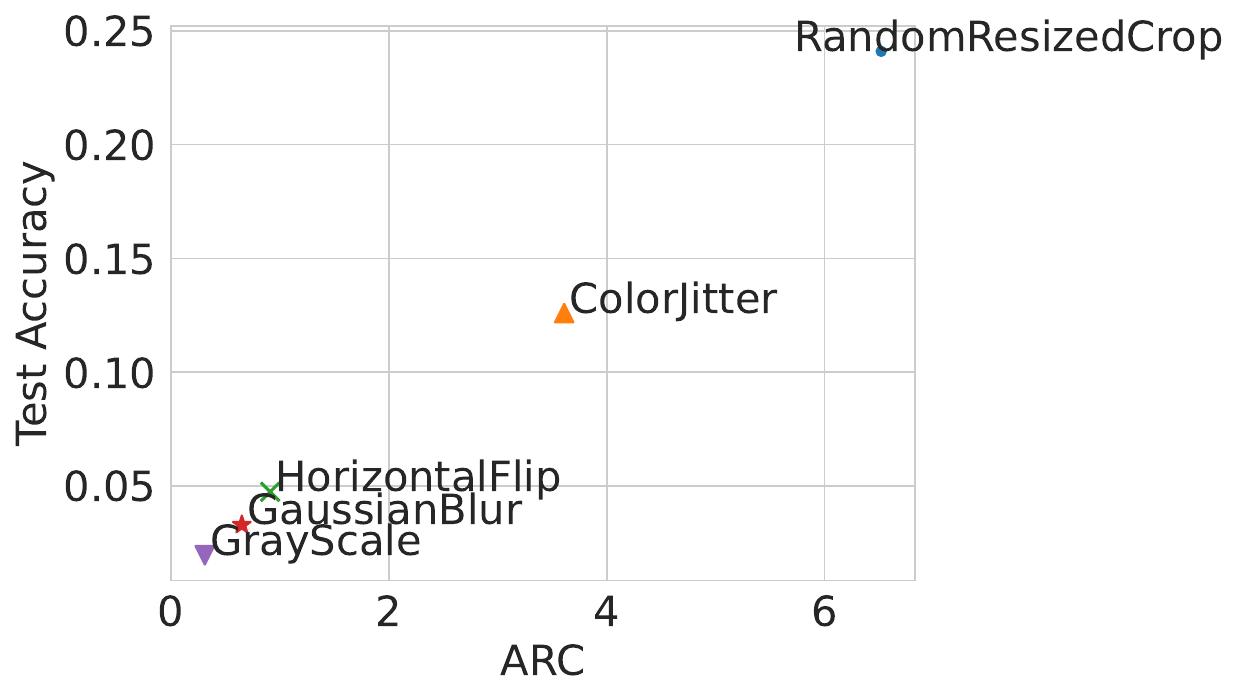}}
    \subfigure[RandomResizedCrop]{
     \includegraphics[width=0.29\textwidth]{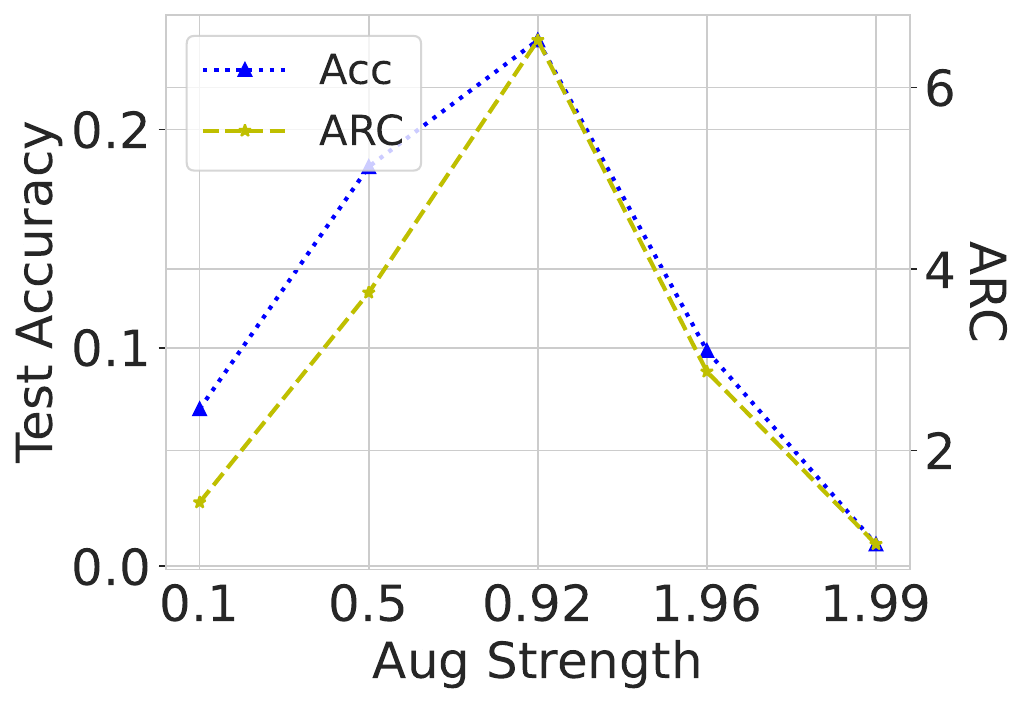}}
         \subfigure[ColorJitter]{
     \includegraphics[width=0.29\textwidth]{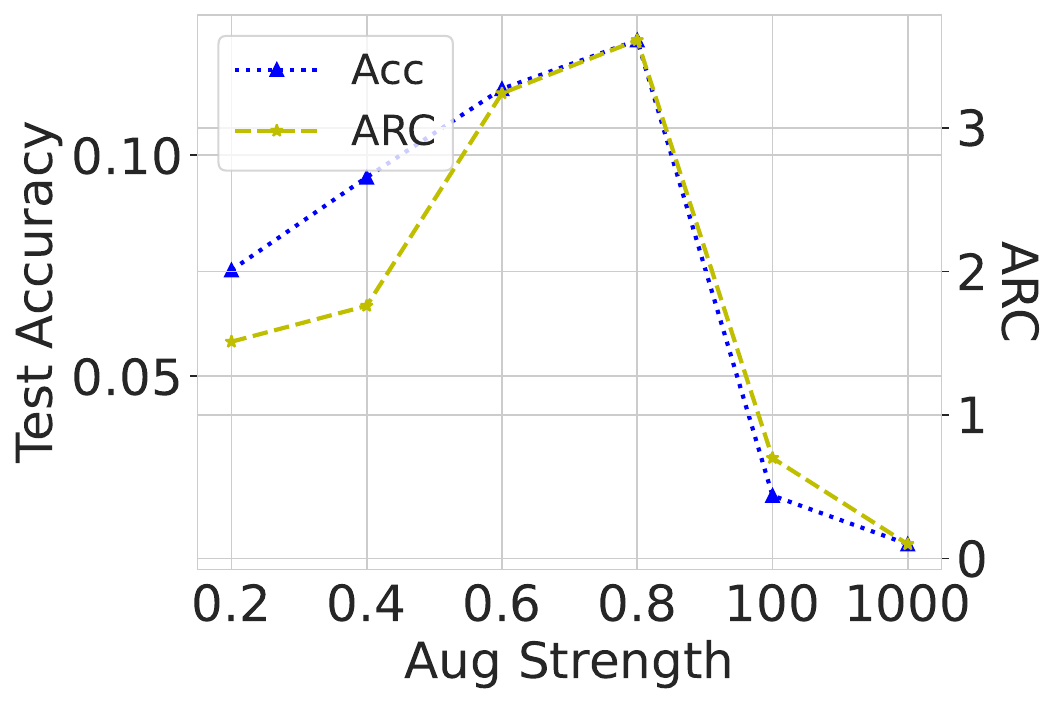}}
    \caption{Average Relative Confusion (ARC) and downstream accuracy \vs different data augmentations adopted in SimCLR with different strengths on ImageNet.}
    \label{fig:ARC-imagenet}
    \vspace{-0.1in}
\end{figure}

\textbf{\textit{ARC on the large-scale data set.}} Besides CIFAR-10, we also evaluate the effectiveness of ARC on the large-scale data set ImageNet. The experiments are conducted on ImageNet with various data augmentations and various strengths including 1) different types of augmentations with default parameters in SimCLR, 2) RandomResizedCrop with different strengths, 3) Colorjitter with different strengths. As shown in Figure \ref{fig:ARC-imagenet}, the proposed ARC aligns well with the linear accuracy, which indicates that the ARC metric also performs well on large-scale data sets.

\subsection{Further Analysis of the Proposed Metric ARC}
\label{sec:garc}

\begin{figure}[!t]
    \centering
    \includegraphics[width=0.8\textwidth]{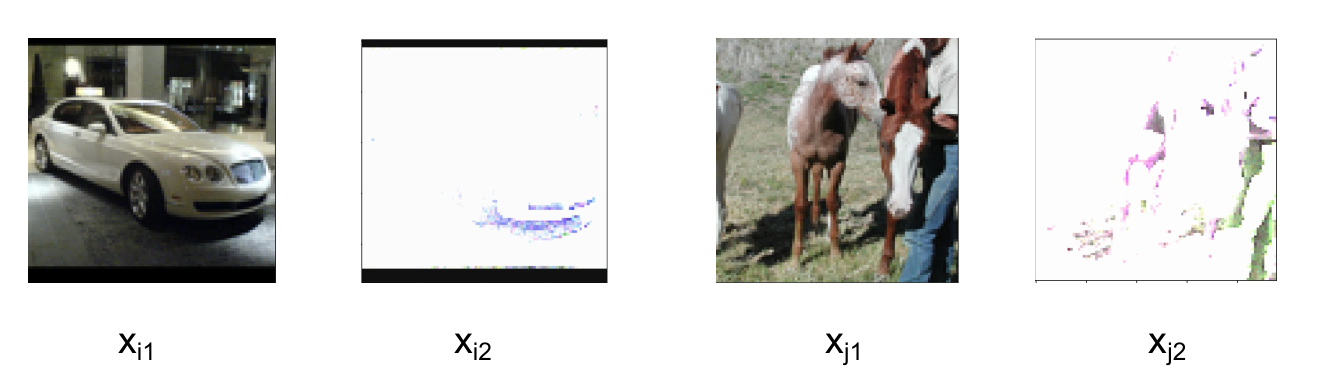}
    \caption{Outliers of data augmentations will mislead the calculation of ARC defined in Section \ref{sec:ARC} $(\|(x_{i2} - x_{i1}\| > \|x_{i2} - x_{j2}\|)$.}
    \label{fig:outlier}
\end{figure}

The proposed metric ARC is based on ACR, while in practice ACR may have outliers to mislead its calculation. For example in Figure \ref{fig:outlier}, $x_{i1}$ and $x_{i2}$ are two augmented views of a car while $x_{j1}$ and $x_{j2}$ are two augmented views of a horse. Due to aggressive augmentations, $x_{i2}$ is more similar to $x_{j2}$ than $x_{i1}$. So the inter-anchor distance $d_{out}(x_{i1})$ is smaller than the intra-anchor distance $d_{in}(x_{i1})$, which results in the wrong increase on ACR, because higher ACR in fact indicates the higher overlap degree while here the car and horse have no semantic overlap.

\textbf{\textit{Generalized ARC.}} To avoid these outliers, we generalize the definition of ACR. For an augmented sample $x_{ij}$, we first compute its distance in the features space to the augmented views of any anchor sample $x_l$: $\gQ(x_{ij}, x_l, f) = \{\|f(x_{ij}) -f(x_{lp}) \|^2, 0\leq p\leq C\}$. Then we respectively use statistic $\gA_1$ to compute the distance of the augmented samples from the same anchor, \ie $d_{in}(x_{ij},\gA_1,f) = \gA_1(\gQ(x_{ij}, x_i, f))$ and  use statistic $\gA_2$ to compute the distance of the augmented samples from different anchors, \ie $d_{out}(x_{ij},\gA_2,f) = \{\gA_2(\gQ(x_{ij}, x_l, f)),l \neq i\}$. Finally we compare $d_{in}(x_{ij},\gA_1,f)$ and the $k$-smallest distance $d_{out-k}(x_{ij},\gA_2,f,k)$ of $d_{out}(x_{ij},\gA_2)$. The generalized ACR is defined as
\begin{equation}
    \operatorname{GACR}(f,\gA_1,\gA_2,k) = \E_{x_{ij} \in \hat{\gD}_u}\mathbb{I}\left(d_{out-k}(x_{ij},\gA_2,f,k) \leq d_{in}(x_{ij},\gA_1,f)\right)
\end{equation}
Similarly, based on the generalized ACR, the generalized ARC is defined as
\begin{equation}
\operatorname{GARC}(\gA_1,\gA_2,k)=\frac{1-\operatorname{GACR}(f_{\rm final},\gA_1,\gA_2,k)}{1-\operatorname{GACR}(f_{\rm init},\gA_1,\gA_2,k)}.
\end{equation}
Note that when $\gA_1$ is set to the maximal value, $\gA_2$ is set to the minimum value, and $k$ is set to 1, GARC degrades to ARC discussed in Section \ref{sec:ARC}. 

\begin{figure}[t]
    \centering
    \subfigure[ARC(max,min)]{
    \label{fig:maxmin}
    \includegraphics[width=.3\textwidth]{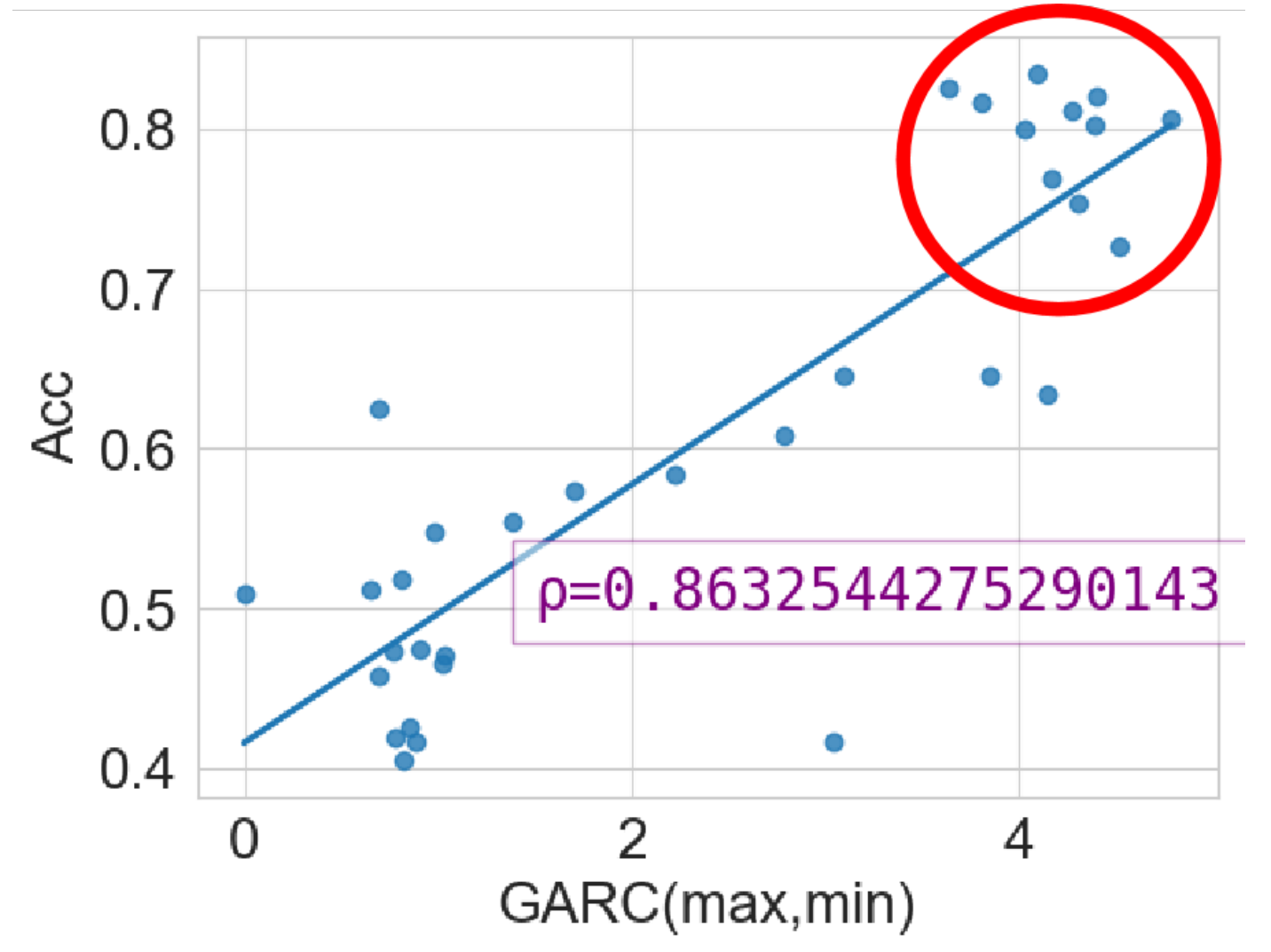}}
    \subfigure[GARC(min,min)]{
    \includegraphics[width=.32\textwidth]{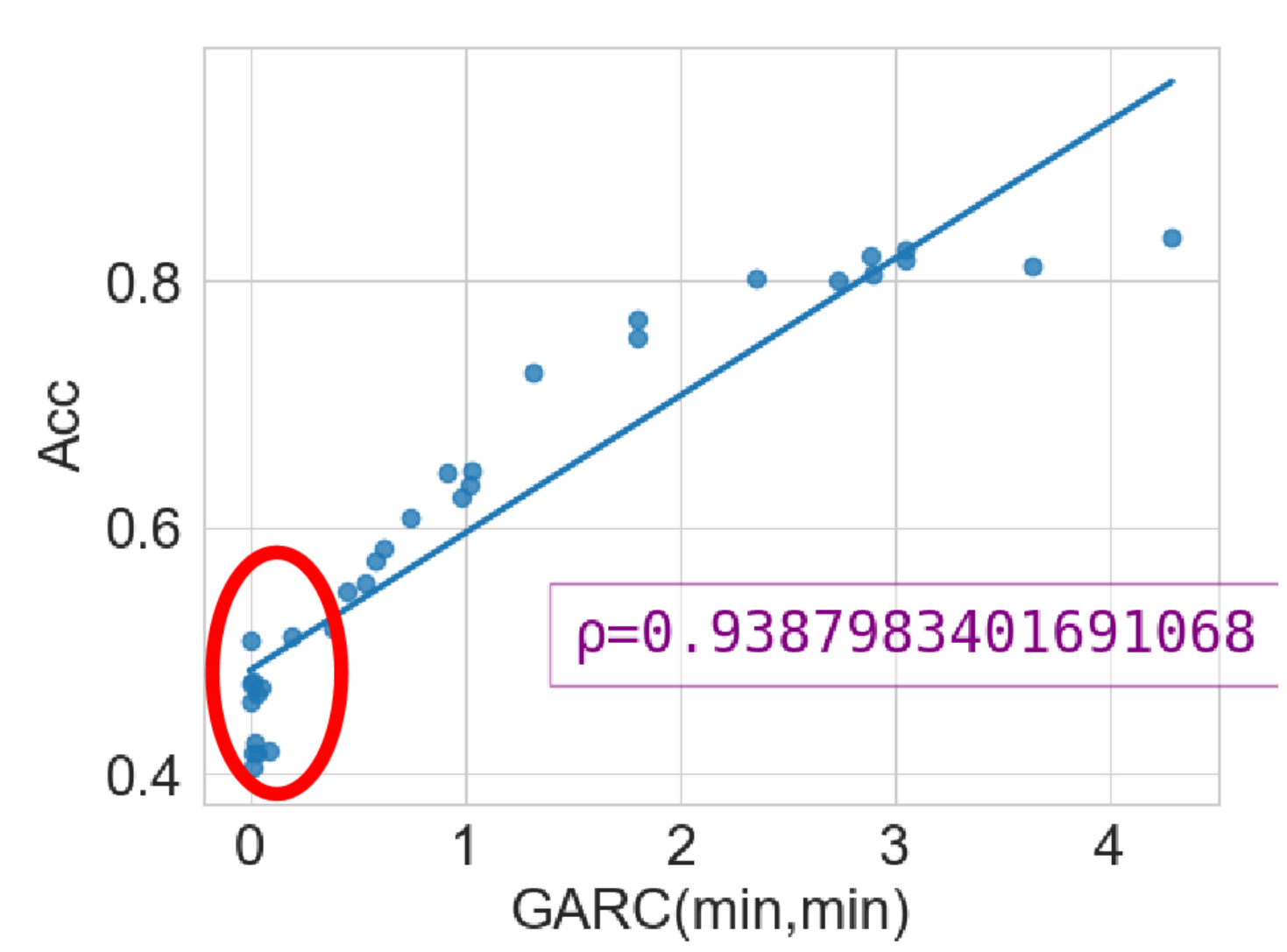}}
    \subfigure[GARC(max,max).]{
    \includegraphics[width=.3\textwidth]{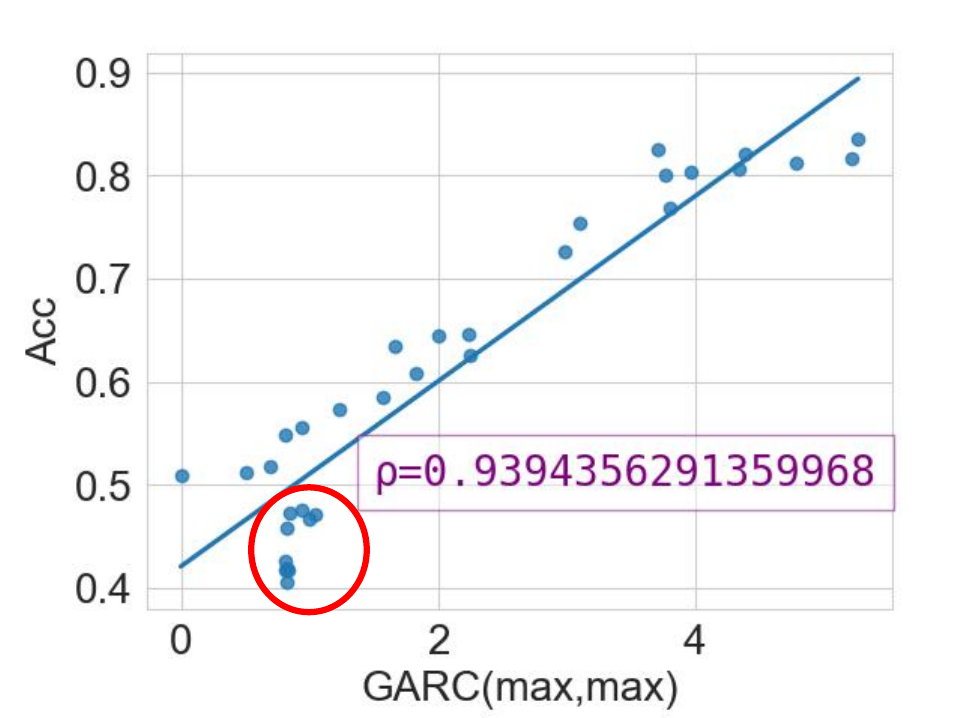}}
    \subfigure[GARC(min,max).]{
    \includegraphics[width=.3\textwidth]{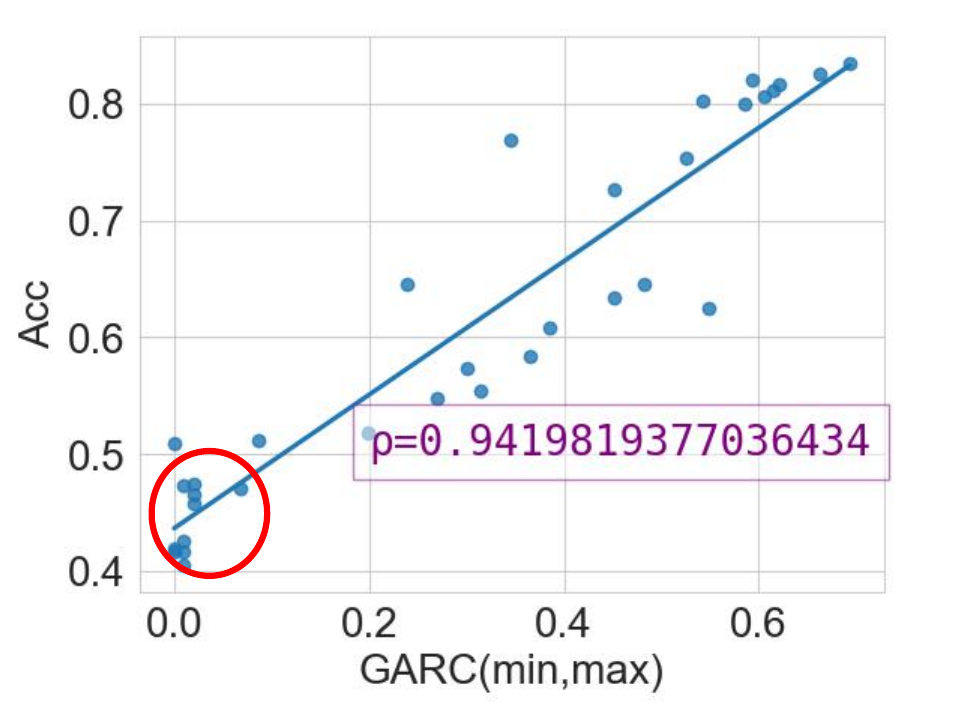}}
        \subfigure[GARC(median,median).]{
    \includegraphics[width=.3\textwidth]{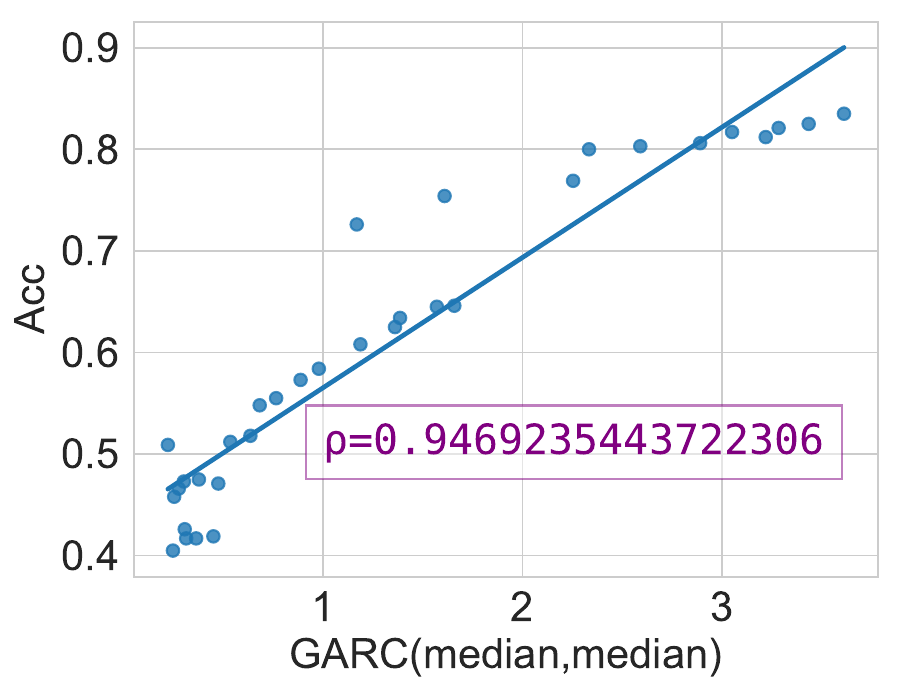}}
        \subfigure[GARC(mean,mean).]{
    \includegraphics[width=.3\textwidth]{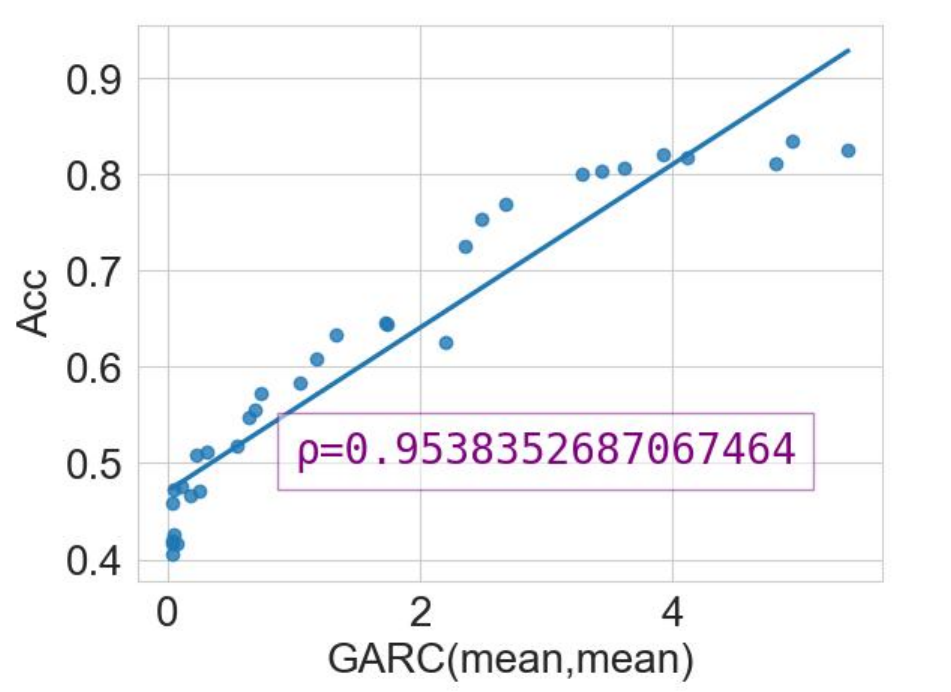}}
    \caption{{The relationship between linear evaluation accuracy and different variants of GARC.}}
    \label{fig:ACR_variant}
\end{figure}

\textbf{\textit{Variants of GARC.}} Following the definition of GARC, here we consider its six different variants, \ie GARC(min, min), GARC(max,max), GARC(max, min), GARC(min, max), GARC(median, median), and GARC(mean, mean). The experiments are conducted on CIFAR-10 with different data augmentations like 1) RandomResizedCrop with different scales of the crop, 2) ColorJitter with different parameters (contrast, brightness, saturation, hue), and 3) composition of all augmentations used in SimCLR with different parameters. The relationship between the downstream accuracy of these models trained with different augmentations and our proposed GARC metric is shown in Figure \ref{fig:ACR_variant}. We find that all the six variants of GARC have a close relationship with the linear accuracy of learned representations. Note that GARC(max,min) is the ARC defined in Section \ref{sec:ARC}. As shown in Figure\ref{fig:maxmin}, the performance of GARC(max,min) decreases when the linear accuracy is high, which is consistent with our findings that aggressive data augmentations will generate outliers and mislead its calculation (marked by red). While GARC(min,min), GARC(max,max), and GARC(min,max) can solve this issue and measures the representations trained with aggressive augmentations more accurately. However, they do not perform well on the representations with low linear accuracy (marked by red). This is because they will ignore some overlapped views, especially with weak data augmentations. In contrast, GARC(median,median) and GARC(mean, mean) can get rid of the influence of these outliers and keeps a strong relationship with the downstream performance of various representations.

\begin{figure}[t]
    \centering
    \subfigure[InfoNCE]{
    \includegraphics[width=.3\textwidth]{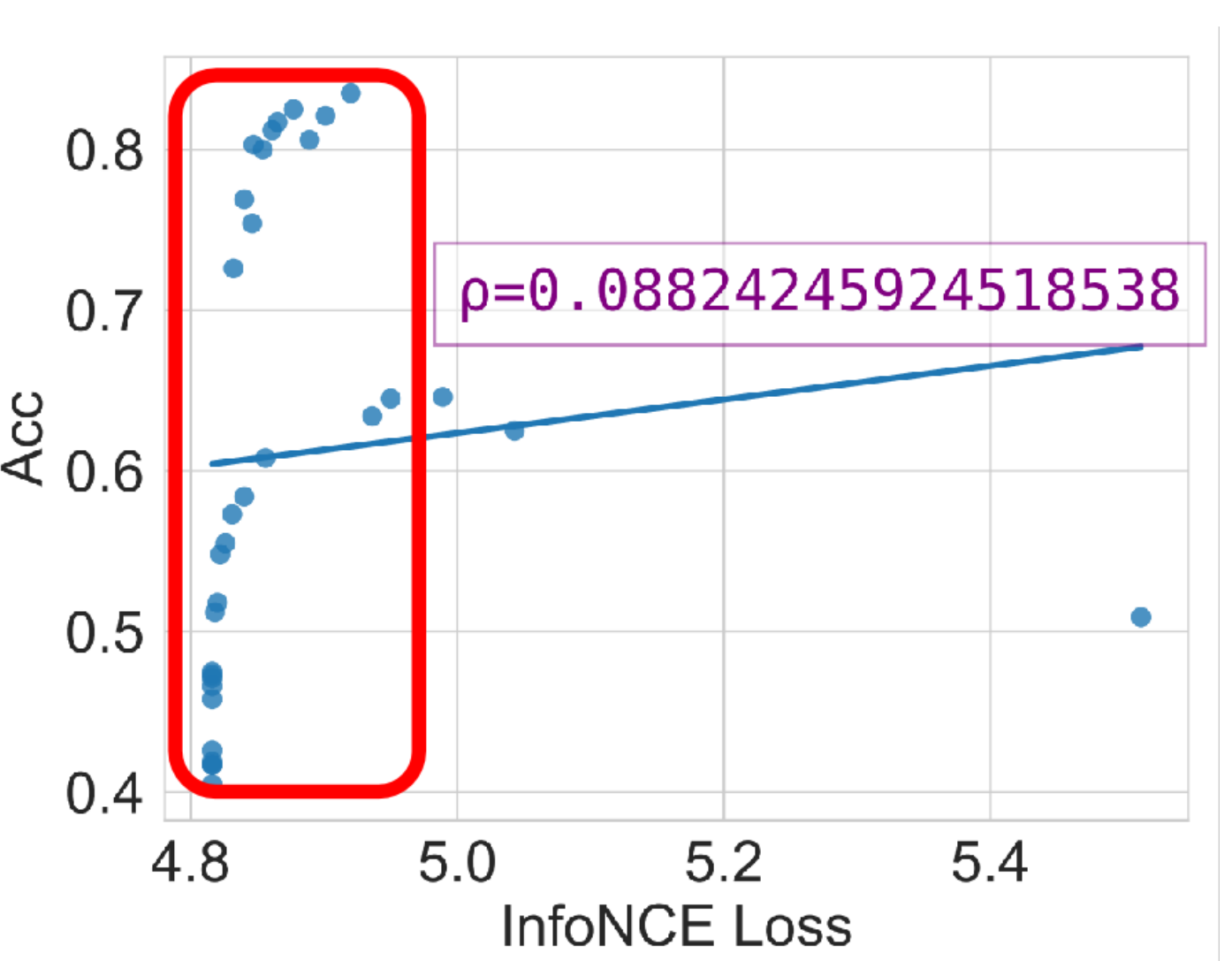}}
    \subfigure[Rotation]{
    \includegraphics[width=.3\textwidth]{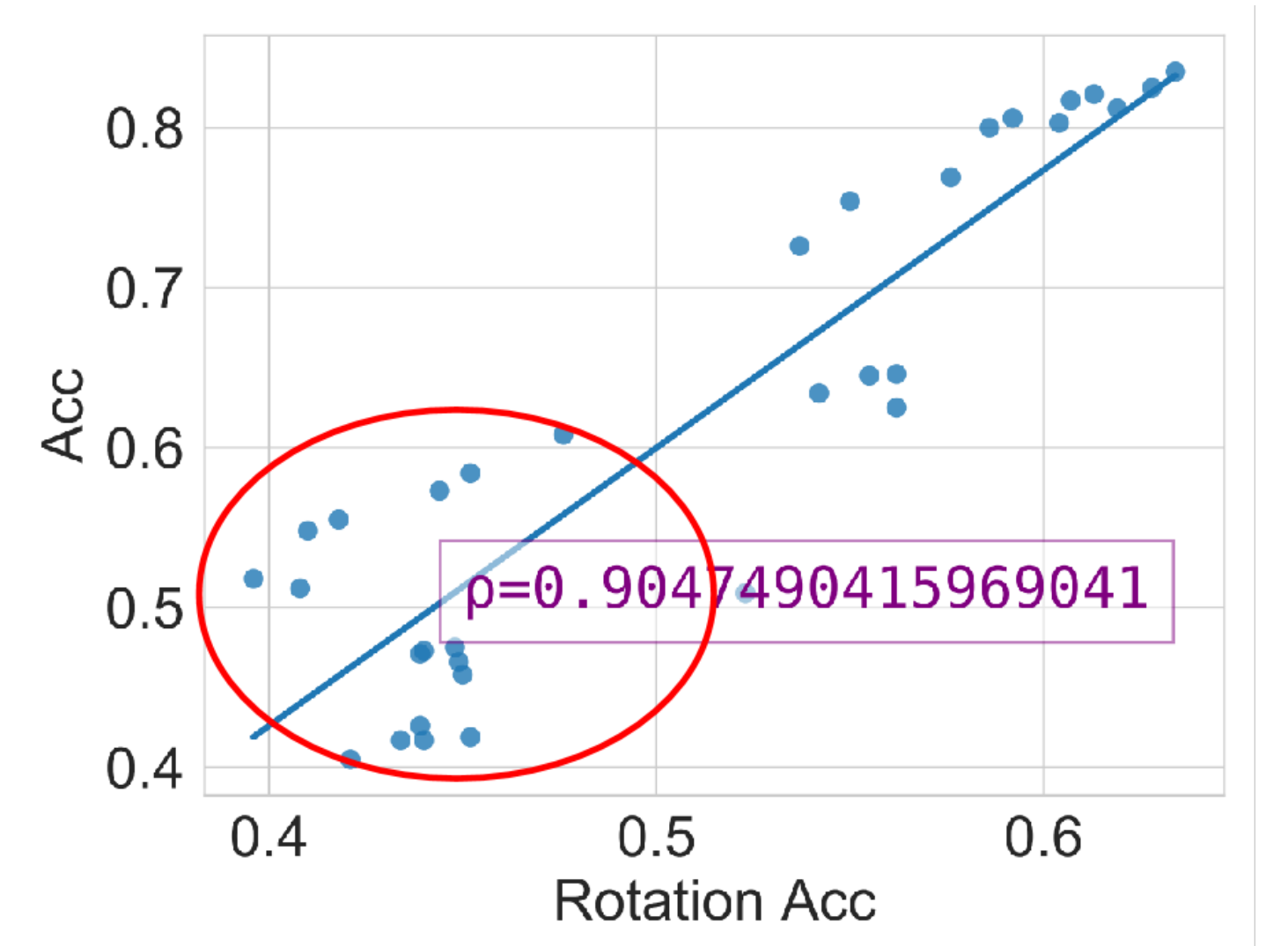}}
    \subfigure[GARC(mean,mean).]{
    \includegraphics[width=.3\textwidth]{figures/MeanMeanLinear.pdf}}
    \caption{{The relationship between linear evaluation accuracy and different unsupervised metrics.}}
    \label{fig:different unsup metric}
\end{figure}

\textbf{\textit{Comparison to other metrics.}}
To further evaluate the performance of our proposed metric, we compare GARC with other unsupervised evaluation metrics such as rotation prediction accuracy \citep{selfaugment}.  They train a linear classifier following the pretrained encoder to predict the rotation angle of the samples and find that the prediction accuracy is strongly related to the downstream classification accuracy. We compare the three different unsupervised metrics, \ie GARC, rotation accuracy, and InfoNCE loss, in Figure \ref{fig:different unsup metric} on CIFAR-10 with the same settings above. {The details can be found in Appendix \ref{sec:different metric}.} 
Figure \ref{fig:different unsup metric} shows that the lower InfoNCE loss does not imply better downstream performance while both our proposed GARC and rotation accuracy are strongly related to the downstream classification accuracy. As the Pearson correlation coefficient $\rho$ \citep{PearsonVIINO} between GARC and linear accuracy is 0.954 while rotation is only 0.905, our proposed metric outperforms rotation accuracy and has a closer relationship to the performance of downstream classification. To be specific, in Figure \ref{fig:different unsup metric} (a), InfoNCE loss does not change a lot with different augmentations (4.8 to 5.0) in most cases while the linear accuracy will increase from 40\% to 80\%, thus InfoNCE is not a good indicator for the downstream performance. Figure \ref{fig:different unsup metric} (b) shows that rotation accuracy performs better in well-clustered representations (right upper corner of the figure) while it performs poorly when the linear accuracy is lower (left lower corner of the figure). Conversely, our proposed GARC keeps a strong relationship with downstream performance of various representations as shown in Figure \ref{fig:different unsup metric} (c). For the computational cost, compared to linear evaluation, rotation accuracy can be obtained without access to supervised labels. However, evaluating rotation accuracy still needs an extra linear classifier and expensive computation costs. In contrast, our proposed GARC directly uses the statistical information of the pretrained models with little additional overheads. In summary, GARC is a well-performed unsupervised model selection metric that is strongly related to the classification performance of the representations and can be obtained with negligible costs.

\section{Conclusion}
\label{sec:conclusion}

In this paper, we proposed a new understanding of contrastive learning through a revisiting of the role of data augmentations. In particular, we notice the aggressive data augmentations applied in contrastive learning can significantly increase the augmentation overlap between intra-class samples, and as a result, by aligning positive samples, we can also cluster inter-class samples together. Based on this insight, we develop a new augmentation overlap theory that could guarantee the downstream performance of contrastive learning without relying on the conditional independence assumption. With this perspective, we also characterize how different augmentation strength affects downstream performance on both random graphs and real-world data sets. Last but not least, we develop a new surrogate metric for evaluating the learned representations of contrastive learning without labels and show that it aligns well with downstream performance. Overall, we believe that we pave a new way for understanding contrastive learning with insights on the designing of contrastive learning methods and evaluation metrics.

\acks{
Yisen Wang was supported by National Key R\&D Program of China (2022ZD0160300), Beijing Natural Science Foundation (L257007), Beijing Major Science and Technology Project under Contract no. Z251100008425006, National Natural Science Foundation of China (92370129, 62376010), Beijing Nova Program (20230484344, 20240484642), and State Key Laboratory of General Artificial Intelligence.}

\appendix

\clearpage

{

}

\section{Proofs}

\subsection{Proof of Lemma \ref{lemma:mc-error}}

\begin{proof}
    First, we have
    \begin{align*}
    &\E_{p(x,z_i)}\left[\log\frac{1}{M}\sum_{i=1}^{M}\exp(f(x)^\top g(z_i)) - \log\E_{p(z_i)}\exp(f(x)^\top g(z_i))\right]\\
    \leq& e\E_{p(x,z_i)}\left[\frac{1}{M}\sum_{i=1}^{M}\exp(f(x)^\top g(z_i))-\E_{p(z_i)}\exp(f(x)^\top g(z_i))\right] \label{Intermediate Value Theorem}=\frac{e}{\sqrt{M}},
    \end{align*}
    where the first inequality follows the Intermediate Value Theorem and $e$ (the natural number) is the upper bound of the absolute derivative of log between two points when $\vert f(x)^\top g(z_i)\vert \leq 1$. And the second inequality uses the following inequality, given the bounded support of $\exp(f(x)^\top g(z_i))$ as following: for i.i.d random variables $Y_i$ with mean $\E Y$ and bounded variance {$\sigma_Y^2 <\sigma^2$,} we have:

{
    \begin{align*}
        &\E\left[\left|\frac{1}{M}\sum_{i=1}^MY_i-\E Y_i\right|\right]\\
        &\leq\sqrt{\E\left[\left|\frac{1}{M}\sum_{i=1}^MY_i-\E Y_i\right|^2\right]}\\
        &=\frac{\sigma}{\sqrt{M}},
    \end{align*}
    }
    where the first inequality follows Jensen's inequality and the second equation follows the property that $Y_i$ is i.i.d random variables. Here, we set $Y_i = \exp(f(x)^\top g(z_i))$. { As $g(z)$ is the mean classifier and $f(x)$ is normalized,  $\vert f(x)^\top g(z_i)\vert \leq 1$,} $\left|Y_i\right| \leq e$. {With Popoviciu's inequality, $Y_i$ has bounded variance $(e)^2$.}\\
    
\end{proof}

\subsection{Proof of Theorem \ref{thm:bounds-CI}}
\label{proof:lower and upper bounds}

We will prove the upper and lower bounds separately as follows.

\subsubsection{The Upper Bound}

We first prove that the mean CE loss can be upper bounded by the InfoNCE loss with Lemma \ref{lemma:mc-error}.
\begin{proof}
Denote $p(x,x^+,y)$ as the joint distribution of the positive pairs $x,x^+$ and the label $y$. Denote the $M$ independently negative smaples as $\{x_i^-\}_{i=1}^M$. Denote $\mu_y$ as the center of features of class $y,\ y=1,\dots,K$. Then we have the following lower bounds of the InfoNCE loss:
{
\begin{align*}
&\bar \gL_{\rm contr}(f)=-\E_{p(x,x^+)}f(x)^\top f(x^+)+\E_{p(x)}\E_{p(x^-_i)}\log \frac{1}{M} \sum_{i=1}^M \exp(f(x)^\top f(x^-_i))\\
\stackrel{(1)}{\geq} & -\E_{p(x,x^+)}f(x)^\top f(x^+)+\E_{p(x)}\log\frac{1}{M}\E_{p(x_i^-)}\sum_{i=1}^M\exp(f(x)^\top f(x^-_i))- \frac{e}{\sqrt{M}}\\
=&-\E_{p(x,x^+)}f(x)^\top f(x^+)+\E_{p(x)}\log\E_{p(x^-)}\exp(f(x)^\top f(x^-))- \frac{e}{\sqrt{M}} \\
=& -\E_{p(x,x^+,y)}f(x)^\top f(x^+)+\E_{p(x)}\log\E_{p(y^-)}\E_{p(x^-|y^-)}\exp(f(x)^\top f(x^-))- \frac{e}{\sqrt{M}}\\
\stackrel{(2)}{\geq}& -\E_{p(x,x^+,y)}f(x)^\top f(x^+)+\E_{p(x)}\log\E_{p(y^-)}\exp(
\E_{p(x^-|y^-)}\left[f(x)^\top f(x^-)\right]) - \frac{e}{\sqrt{M}}\\
\stackrel{(3)}{=}& -\E_{p(x,y)}f(x)^\top\mu_y+\E_{p(x)}\log\E_{p(y^-)}\exp(\E_{p(x^-|y^-)}\left[f(x)^\top f(x^-)\right])- \frac{e}{\sqrt{M}}\\
{=} &-\E_{p(x,y)}f(x)^\top\mu_{y}  +\E_{p(x)}\log\E_{p(y^-)}\exp(f(x)^\top\mu_{y^-})- \frac{e}{\sqrt{M}}\\
=&-\E_{p(x,y)}f(x)^\top\mu_y +\E_{p(x)} \log\frac{1}{K}\sum_{k=1}^K\exp(f(x)^\top\mu_k) - \frac{e}{\sqrt{M}} \\
=&\E_{p(x,y)}\big[-f(x)^\top\mu_y+\log \frac{1}{K} \sum_{k=1}^K\exp(f(x)^\top\mu_k)\big]- \frac{e}{\sqrt{M}}  \\
=& \bar \gL_{\rm mCE}(f) - \frac{e}{\sqrt{M}} ,
\end{align*}
}
which is equivalent to our desired results. In the proof above,
(1) follows Lemma \ref{lemma:mc-error}; (2) follows the Jensen's inequality for the convex function $\exp(\cdot)$; (3) follows the conditional independence assumption. 
\end{proof}

\subsubsection{The Lower Bound}
In this part, we further show a lower bound on the downstream performance. 
{
\begin{Lemma}[\cite{reverse_jensen} Corollary 3.5 (restated)]
\label{lemma:reverse-jensen}
Let $g:\mathbb{R}^m \rightarrow \mathbb{R}$ be a differentiable convex mapping and $z \in \mathbb{R}^m$. Suppose that $g$ is $L$- smooth with the constant $L > 0$, \ie $\forall x,y\in\sR^m, \Vert \nabla g(x) - \nabla g(y) \Vert \leq L \Vert x-y\Vert$. Then we have
\begin{equation}
\begin{aligned}
0 &\leq \E_{p(z)}g(z) - g\left(\E_{p(z)}z\right)\\
&\leq L\left[ \E_{p(z)} \lVert z\rVert^2 - \Vert \E_{p(z)} z\Vert^2\right]\\
&= L[\sum\limits_{j=1}^m \E_{p(z)}\Vert z^{(j)}\Vert^2 - \sum\limits_{j=1}^m \Vert\E_{p(z)} z^{(j)}\Vert^2 ]\\
&= L[\sum\limits_{j=1}^m \E_{p(z^{(j)})}\Vert z^{(j)}\Vert^2 - \sum\limits_{j=1}^m \Vert\E_{p(z^{(j)})} z^{(j)}\Vert^2 ]\\
&= L \sum_{j=1}^{m}\var(z^{(j)})  
\end{aligned}
\end{equation}
where $x^{(j)}$ denotes the $j$-th dimension of $x$.
\label{lemma:reverse jensen}
\end{Lemma}
}
With the lemma above, we can derive the lower bound of the downstream performance.
\begin{proof}
Similar to the proof of the upper bound, we have
\begin{align*}
&\bar \gL_{\rm mCE}(f)=-\E_{p(x,y)}f(x)^\top\mu_y +\E_{p(x)} \log \frac{1}{K} \sum_{i=1}^K\exp(f(x)^\top\mu_i) \\
=&-\E_{p(x,y)}f(x)^\top\mu_y +\E_{p(x)} \log \frac{1}{K}\sum_{i=1}^K\exp(f(x)^\top\mu_i) \\
=&-\E_{p(x,y)}f(x)^\top\mu_y + \E_{p(x)}\log\E_{p(y^-_i)}\exp(f(x)^\top\mu_{y_i}) \\
\stackrel{(1)}{\geq}& -\E_{p(x,y)}f(x)^\top\mu_y+ \E_{p(x)}\E_{p(y^-_i)}\log\frac{1}{M}\sum_{i=1}^M\exp(f(x)^\top \mu_{y_i}) -\frac{e}{\sqrt{M}} \\
\stackrel{(2)}{=}&-\E_{p(x,x^+)}f(x)^\top f(x^+)  
+\E_{p(x)}\E_{p(y^-_i)}\log\frac{1}{M}\sum_{i=1}^M\exp(\E_{p(x^-_i|y^-_i)}f(x)^\top f(x^-_i)) -\frac{e}{\sqrt{M}} \\
\stackrel{(3)}{\geq}&-\E_{p(x,x^+)}f(x)^\top f(x^+) \\
&+\E_{p(x)}\E_{p(y^-_i)}\E_{p(x^-_i|y)}\left[\log\frac{1}{M}\sum_{i=1}^M\exp(f(x)^\top f(x^-))\right]- \frac{1}{2}\sum_{j=1}^m\var(f_j(x^-)\mid y) -\frac{e}{\sqrt{M}} \\
{=}&-\E_{p(x,x^+)}\left[f(x)^\top f(x^+)+\E_{p(x^-_i)}\log\sum_{i=1}^M\exp(f(x)^\top f(x^-))\right]\\
&\quad\quad\quad\quad\quad\quad
- \frac{1}{2}\sum_{j=1}^m\var(f_j(x^-)\mid y) -\frac{e}{\sqrt{M}} \\
=& \bar \gL_{\rm contr}(f)- \frac{1}{2}\var(f(x)\mid y) -\frac{e}{\sqrt{M}},
\end{align*}
which is our desired result. In the proof, (1) we adopt the Monte Carlo estimate with $M$ samples from $p(y)$ and bound the approximation error with Lemma \ref{lemma:mc-error}; (2) follows the conditional independence assumption; (3) we first show that the convex function $\operatorname{logsumexp}$ is $L$-smooth as a function of $f(x^-_j)$ in our scenario. Because $\Vert f(X) \Vert\leq 1$,  we have $\forall f(x_{j_1}), f(x_{j_2})\in\sR^m$, the following bound on the difference of their gradients holds
\begin{align*}
&\left\|\frac{\partial \log[\exp(f(x)^\top f(x_{j_1}^-)+\sum_{i\neq j}\exp(f(x)^\top f(x_i^-)) )]}{\partial f(x_{j_1}^-)} - \frac{\partial \log[\exp(f(x)^\top f(x_{j_2}^-)+\sum_{i\neq j}\exp(f(x)^\top f(x_i^-)) )]}{\partial f(x_{j_2}^-)}\right\|\\
=&\left\|\left(\frac{\exp(f(x)^\top f(x_{j_1}^-))}{\exp(f(x)^\top f(x_{j_1}^-)+\sum_{i\neq j}\exp( f(x_i^-)) )}-\frac{\exp(f(x)^\top f(x_{j_2}^-))}{\exp(f(x)^\top f(x_{j_2}^-)+\sum_{i\neq j}\exp(f(x)^\top f(x_i^-)) )}\right)f(x)\right\|\\
\leq&\left|\frac{(\sum_{i\neq j}\exp(f(x)^\top f(x_i^-)) \exp( f(x_{j_1}^-))-\sum_{i\neq j}\exp(f(x)^\top f(x_i^-)) \exp(f(x)^\top f(x_{j_2}^-))}{(\exp(f(x)^\top f(x_{j_1}^-))+\sum_{i\neq j}\exp(f(x)^\top f(x_i^-)) )(\exp(f(x)^\top f(x_{j_2}^-))+\sum_{i\neq j}\exp(f(x)^\top f(x_i^-)) )}\right|\cdot\left\|f(x)\right\|\\
\leq& \|f(x)\| \leq \frac{1}{2}\left\|f(x_{j_1}^-)-f(x_{j_2}^-)\right\|. 
\end{align*}
So here the $\operatorname{logsumexp}$ is $L$-smooth for $L=\frac{1}{2}$. Then, we can apply the reversed Jensen's inequality in Lemma \ref{lemma:reverse-jensen}.

\end{proof}

{
\subsection{Proof of Theorem \ref{thm:general-generalization-gap} with Labeling Error}

Similar to Theorem \ref{thm:bounds-CI}, we will prove the upper and lower bounds separately as follows.

\subsubsection{The Upper Bound}

With Lemma \ref{lemma:mc-error}, we show that upper bounds the mean CE loss with the InfoNCE loss.
\begin{proof}
We denote $p(x,x^+,y)$ as the joint distribution of the positive pairs $x,x^+$ and the label $y$ of $x$. Denote the $M$ independently negative smaples as $\{x_i^-\}_{i=1}^M$. Denote $\mu_y$ as the center of features of class $y,\ y=1,\dots,K$. Similar to the proof of Theorem \ref{thm:bounds-CI}, we have
\begin{align*}
&\bar\gL_{\rm contr}(f)=-\E_{p(x,x^+)}f(x)^\top f(x^+)+\E_{p(x)}\E_{p(x^-_i)}\log \frac{1}{M} \sum_{i=1}^M\exp(f(x)^\top f(x^-_i))\\
\stackrel{(1)}{\geq}&-\E_{p(x,x^+,y)}f(x)^\top(\mu_{y} + f(x^+)-\mu_y) +\E_{p(x)}\log\E_{p(y^-)}\exp(\E_{p(x^-|y^-)}\left[f(x)^\top f(x^-)\right])- \frac{e}{\sqrt{M}}\\
{=} &-\E_{p(x,x^+,y)}[f(x)^\top\mu_{y} + f(x)^\top(f(x^+)-\mu_y)] +\E_{p(x)}\log\E_{p(y^-)}\exp(f(x)^\top\mu_{y^-})- \frac{e}{\sqrt{M}}\\
\stackrel{(2)}{\geq} &-\E_{p(x,x^+,y)}\left[f(x)^\top\mu_{y} + \Vert(f(x^+)-\mu_y)\Vert\right] +\E_{p(x)}\log\E_{p(y^-)}\exp(f(x)^\top\mu_{y^-})- \frac{e}{\sqrt{M}}\\
\stackrel{(3)}{\geq} &-\E_{p(x,y)}f(x)^\top\mu_{y} -\sqrt{\E_{p(x,y)}\Vert f(x^+)-\mu_y\Vert^2} +\E_{p(x)}\log\E_{p(y^-)}\exp(f(x)^\top\mu_{y^-})- \frac{e}{\sqrt{M}}\\
=&-\E_{p(x,y)}f(x)^\top\mu_y -\sqrt{\E_{p(x,y)}\Vert f(x^+)-\mu_{y^+} + \mu_{y^+} - \mu_{y} \Vert^2} 
+\E_{p(x)} \log\frac{1}{K}\sum_{k=1}^K\exp(f(x)^\top\mu_k) - \frac{e}{\sqrt{M}}\\
\geq &\E_{p(x,y)}\big[-f(x)^\top\mu_y+\log \frac{1}{K} \sum_{k=1}^K\exp(f(x)^\top\mu_k)\big]-2\sqrt{\var( f(x^+)\mid y(x^+))}- 2\sqrt{\Vert \mu_{y^+} - \mu_y\Vert ^2}- \frac{e}{\sqrt{M}}  \\
\stackrel{(4)}{\geq} &\E_{p(x,y)}\big[-f(x)^\top\mu_y+\log \frac{1}{K} \sum_{k=1}^K\exp(f(x)^\top\mu_k)\big]-2\sqrt{\var( f(x^+)\mid y(x^+))} - 4\sqrt{\alpha}- \frac{e}{\sqrt{M}}  \\
=&\bar \gL_{\rm mCE}(f) -2\sqrt{\var( f(x^+)\mid y^+)} - 4\sqrt{\alpha}- \frac{e}{\sqrt{M}}\\
=&\bar \gL_{\rm mCE}(f) -2\sqrt{\var( f(x)\mid y)} - 4\sqrt{\alpha}- \frac{e}{\sqrt{M}},
\end{align*}
which is equivalent to our desired results. In the proof above, (1) follows Lemma \ref{lemma:mc-error}, (2) follows from the fact that because $f(x)\in\sS^{m-1}$, we have
\begin{equation}
    f(x)^\top(f(x^+)-\mu_y)\leq \left(\frac{f(x^+)-\mu_y}{\Vert f(x^+)-\mu_y\Vert}\right)^\top(f(x^+)-\mu_y)=\Vert f(x^+)-\mu_y\Vert;
\end{equation}
(3) follows the Jensen inequality and the fact that because $p(x,x^+)=p(x^+,x)$ holds, $x,x^+$ have the same marginal distribution and (4) follows assumption \ref{assumption:label-consistency} and $f(x)$ is normalized. We note that when we throw the conditional independence assumption, we do not have $\E_{p(x,x^+)}f(x)^\top = \E_{p(x,y)}f(x)^\top \mu_y$ and there exists an additional variance term $\sqrt{\var( f(x^+)\mid y^+)}$.
\end{proof}

\subsubsection{The Lower Bound}

With the Lemma \ref{lemma:reverse jensen}, we can derive the lower bound of the downstream performance.

\begin{proof}
Similar to the proof of the upper bound, we have
\begin{align*}
&\bar\gL_{\rm mCE}(f)=-\E_{p(x,y)}f(x)^\top\mu_y +\E_{p(x)} \log\frac{1}{K}\sum_{i=1}^K\exp(f(x)^\top\mu_i) \\
=&-\E_{p(x,y)}f(x)^\top\mu_y + \E_{p(x)}\log\E_{p(y^-_i)}\exp(f(x)^\top\mu_{y_i})\\
\stackrel{(1)}{\geq}&-\E_{p(x,y)}[f(x)^\top f(x^+) + f(x)^\top(\mu_y -f(x^+))] + \E_{p(x)}\E_{p(y^-_i)}\log\frac{1}{M}\sum_{i=1}^M\exp(f(x)^\top \mu_{y_i}) -\frac{e}{\sqrt{M}} \\
\stackrel{(2)}{\geq}&-\E_{p(x,x^+)}f(x)^\top f(x^+) -\E_{p(x^+,y)}\Vert f(x^+)^\top -\mu_y\Vert \\
&\quad\quad\quad\quad\quad\quad
+\E_{p(x)}\E_{p(y^-_i)}\log\frac{1}{M}\sum_{i=1}^M\exp(\E_{p(x^-_i|y^-_i)}f(x)^\top f(x^-_i)) -\frac{e}{\sqrt{M}} \\
\stackrel{(3)}{\geq}&-\E_{p(x,x^+)}f(x)^\top f(x^+)-2\sqrt{\var(f(x^+)\mid y^+)} -4\sqrt{\alpha}\\
&+\E_{p(x)}\E_{p(y^-_i)}\E_{p(x^-_i|y)}\left[\log\frac{1}{M}\sum_{i=1}^M\exp(f(x)^\top f(x^-))\right]- \frac{1}{2}\sum_{j=1}^m\var(f_j(x^-)\mid y) -\frac{e}{\sqrt{M}} \\
=&\bar \gL_{\rm contr}(f)-2\sqrt{\var(f(x)\mid y)}- \frac{1}{2}\var(f(x)\mid y) -\frac{e}{\sqrt{M}}--4\sqrt{\alpha},
\end{align*}
which is our desired result. In the proof, (1) we adopt the Monte Carlo estimate with $M$ samples from $p(y)$ and bound the approximation error with Lemma \ref{lemma:mc-error}; (2) follows the same deduction in the upper bound; {(3) the second term is derived following the Jensen inequality for the alignment term. As for the third term,} we prove that the convex function $\operatorname{logsumexp}$ is $L$-smooth as a function of $f(x^-_j)$ in the proof of Theorem \ref{thm:bounds-CI}. Then, we can apply the reversed Jensen's inequality in Lemma \ref{lemma:reverse-jensen}. Similar to the upper bound, when we throw conditional independence assumption, the additional variance term $\sqrt{\var(f(x)\mid y)}$ arises.

\end{proof}

}

\subsection{Proof of Proposition \ref{prop:wang-couterexample}}
\label{proof of wang counter}

\begin{proof}
{We only need to give a counterexample that satisfies the desired classification accuracy. We consider the case where any pair of samples from $\{x_i\}_{i=1}^N$ will not be aligned, which is easily achieved if we adopt a small enough data augmentation. 
In this scenario, 
the perfect alignment of positive samples $(x_i, x^+_i)$ could have no effect on the other samples. 
Therefore, when
the features  $\{f(x_i)\}_{i=1}^N$ are uniformly distributed in $\sS^{m-1}$, according to the law of large number, for any measurable set $\gU\in\sS^{m-1}$, when $N$ is large enough, there will be almost equal size of features from each class in $\gU$. Consequently, any classifier $g$ that classifies $\gU$ to class $k$ will only have $1/K$ accuracy asymptotically.}
\end{proof} 

\subsection{Proof of Theorem \ref{thm:weak-generalization-gap}}

\begin{proof} 
Consider any pair of samples $(x,x')$ from the same class $y$, and the positive sample of $x$ as $x^+$. As intra-class connectivity holds, $x$ and $x'$ are connected, and the maximal length of the path from $x$ to $x'$ is $D$. Therefore, we can bound the representation distance between $x$ and $x'$ by the triangular inequality under the $\varepsilon$-alignment 
\begin{equation}
\Vert f(x) - f(x') \Vert {\leq} D\sup_{(x,x^+)\sim p(x,x^+)} \Vert f(x) - f(x_+)\Vert {\leq} D\varepsilon 
\label{eq:any-pair-bound}
\end{equation}
With the inequality above, we can bound the variance terms in Theorem \ref{thm:general-generalization-gap}. In particular, the conditional variance can be bounded as
\begin{equation}
\begin{aligned}
&\Var(f(x) \mid y)\\
{=}& \E_{p(y)}\E_{p(x|y)}\Vert f(x) - \E_{x'}f(x')\Vert^2\\
{\leq}& \E_{p(y)}{\E_{p(x|y)}\E_{p(x'|y)}\Vert f(x) - f(x')\Vert^2}\\
{\leq}& \E_{p(y)}\max_{x,x'\sim p(x|y)} \Vert f(x) - f(x')\Vert^2\\
\stackrel{(1)}{\leq}& \E_{p(y)}{D^2\varepsilon^2}=D^2\varepsilon^2,
\end{aligned}
\label{eq:var-bound}
\end{equation}
where (1) follows Eq.~\ref{eq:any-pair-bound}. 
Then, we can bound the variance items in Theorem \ref{thm:general-generalization-gap} with Eq.~\ref{eq:var-bound}.
\end{proof}

\subsection{Proof of Corollary \ref{thm:closed-generalization-gap}}

\begin{proof}
Combing with Theorem \ref{thm:weak-generalization-gap} and $\epsilon =0 $, the intra-class variance term varnishes. So we can directly obtain this result.
\end{proof}
\subsection{Proof of Corollary \ref{cor:spectral}}

\begin{proof}
$\mu_{1k},\mu_{2k},\cdots$ are orthonormal eigenvectors of adjacent matrix of the intra-class graph $\gG_k$ with eigenvalues $\lambda_{1k},\lambda_{2k},\cdots$ $( \lvert \lambda_{1k} \rvert \geq \lvert \lambda_{2k} \rvert \geq \cdots)$. Let $M$ denote the adjacency matrix of graph $\gG_k$ and $(M)_{r,s}$ denote the row $r$ and column $s$ element of $M$, we have $M = \sum\limits_i \lambda_{ik}\mu_{ik}\mu_{ik}^\top$
\begin{equation}
\begin{aligned}
(M^q)_{r,s} &= \sum\limits_i \lambda_{ik}^q (\mu_{ik}\mu_{ik}^\top)_{r,s}\\
&= \lambda_{1k}^q (\mu_{1k}\mu_{1k}^\top)_{r,s}+\sum\limits_{i>1} \lambda_{ik}^q (\mu_{ik}\mu_{ik}^\top)_{r,s}\\
&\geq \lvert\lambda_{1k}\lvert^q \omega_k^2 - \lvert \lambda_{2k}\rvert ^q\sum\limits_{i>1}\lvert(\mu_{ik}\mu_{ik}^\top)_{r,s}\rvert \\
&=\lvert\lambda_{1k}\lvert^q \omega_k^2 - \lvert \lambda_{2k}\rvert ^q\sum\limits_{i>1}\lvert(\mu_{ik})_r\rvert\lvert(\mu_{ik})_s\rvert\\
&\stackrel{(1)}{\geq}\lvert\lambda_{1k}\lvert^q \omega_k^2 - \lvert \lambda_{2k}\rvert ^q\left(\sum\limits_{i>1}\lvert(\mu_{ik})_r\rvert^2\right)^{\frac{1}{2}} \left(\sum\limits_{i>1}\lvert(\mu_{ik})_s\rvert^2\right)^{\frac{1}{2}}\\
&=\lvert\lambda_{1k}\lvert^q \omega_k^2 - \lvert \lambda_{2k}\rvert ^m\left(1-(u_{1k})_r^2\right)^{\frac{1}{2}} \left(1-(u_{1k})_s^2\right)^{\frac{1}{2}}\\
&\geq \lvert\lambda_{1k}\lvert^q \omega_k^2 - \lvert \lambda_{2k}\rvert ^q\left(1-\omega_k^2\right),
\end{aligned}
\end{equation}
where (1) employs the Cauchy–Schwarz inequality.
So, if $q\geq\frac{\log((1-\omega_k^2)/\omega_k^2)}{\log(\lvert \lambda_{1k} \rvert /\lvert \lambda_{2k} \rvert )}$, every element of $(M)^q>0$, \ie$D_k \leq \frac{\log((1-\omega_k^2)/\omega_k^2)}{\log(\lvert \lambda_{1k} \rvert /\lvert \lambda_{2k} \rvert )}$. For the maximal radius $D$, we have $D \leq \frac{\log((1-\omega^2)/\omega^2)}{\log(\lvert \lambda_1 \rvert /\lvert \lambda_2 \rvert )}$, where $\omega =  {\rm min}_{i,k}\lvert (\mu_{1k})_i \rvert, \lambda_1 =  {\rm min}_k \vert\lambda_{1k}\vert, \lambda_2 =  {\rm max}_k \vert\lambda_{2k}\vert$.
\end{proof}
With this Lemma, we can directly obtain corollary \ref{cor:spectral}.

\subsection{Proof of Cases That HaoChen et al. (2021) Fail to Analyze} 

When Assumption \ref{ass:intra-class connectivity} holds, the error of labeling function $\alpha = 0$.
\label{pro:fail calse 1}

\textit{Case I:} If Assumption \ref{ass:intra-class connectivity} holds and $\lfloor m/2 \rfloor \leq K$, then $\rho_{\lfloor m/2 \rfloor} = 0$. When Assumption \ref{assumption:label-consistency} holds, then $\alpha = 0$. The bound in Lemma \ref{thm:spectral bound} becomes a $\frac{0}{0}$ term which fails to be analyzed.

\begin{proof}
We give a solution of the sparsest partition of augmentation graph. $\forall t \leq \lfloor m/2 \rfloor-1$, we set $S_t = \gG_t$, where $\gG_t$ is the subgraph of latent class $t$, and $S_{\lfloor m/2 \rfloor} = \gG - (S_1 \bigcap S_2 \bigcap \cdots \bigcap S_{\lfloor m/2 \rfloor-1})$. As Assumption \ref{ass:intra-class connectivity} holds, there exists no edge between different latent classes, \ie $\phi_\gG(S_1) = \phi_\gG(S_1) = \cdots = \phi_\gG(S_{\lfloor m/2 \rfloor-1}) = 0 $, where $\phi$ is Dirichlet conductance of augmentation graph. So we have $\rho_{\lfloor m/2 \rfloor} = 0$.
\end{proof}

\textit{Case II:} If we adopt the same settings in Proposition \ref{prop:wang-couterexample}, \ie there exists no support overlapping, the bound in Lemma \ref{thm:spectral bound} becomes a $\frac{0}{0}$ term again.

\begin{proof}
When there exist no overlap between intra-class samples as analyzed in Proposition \ref{prop:wang-couterexample}, there exist no edge between different intra-class samples. So $\phi_\gG(S_1) = \phi_\gG(S_1) = \cdots = \phi_\gG(S_{\lfloor m/2 \rfloor-1}) = 0$. So we have $\rho_{\lfloor m/2 \rfloor} = 0$.
\end{proof}

\subsection{Proof of Theorem \ref{theorem:augmentation-strength}}

\begin{proof}
From definition and notation in section 5. We can construct an augmentation Graph $\gG(V,E(\gT))$ given N random samples. We define $D_k$ as the distance from a random point to its k-th nearest neighbour. \cite{PERCUS1998424} discuss $D_k$ in random grpah and give the estimation of that: 
 \begin{equation}
     \begin{aligned}
     D_k \approx \frac{[(d/2)!]^{\frac{1}{d}}}{\sqrt{\pi}}\frac{(k-1+1/d)!}{(k-1)!}(\frac{S}{N-1})^{\frac{1}{d}}[1-\frac{1/d+1/d^2}{2(N-1)}+O(\frac{1}{(N-1)^2})],
     \end{aligned}
 \end{equation}
 where d is the dimension of hypersphere, S is the surface area of sample distribution and N is the number of random points. When $r < D_1$ there is no edge in the graph. So the class is separated. When $r > D_{N-1}$, any pair of vertexes have an edge between them, so the graph is full connected. With this conclusion, we can directly have Theorem \ref{theorem:augmentation-strength}.
 \end{proof}
 
\subsection{Proof of Theorem \ref{theorem:augmetantion-infinity}}

 \begin{proof}
Denote
 \begin{equation}
 \begin{aligned}
     r_{mc} = \inf \{r_i>0:G_N(V,E,r_i) \text{is connected}\}.
 \end{aligned}
 \end{equation}

 With Theorem 1.1 from \cite{10.1214/aop/1022677261} and features are uniformly distributed in the surface of unit hypersphere, we have
 \begin{equation}
 \begin{aligned}
 \lim\limits_{N \to \infty}(r_{mc}^d\frac{N^2}{\log N})= 2\frac{(1-\frac{1}{d})S}{V_u}, d\geq 2,
 \end{aligned}
 \end{equation}
 where $V_u$ denotes to the volume of unit hypershpere.
\end{proof}

\section{Additional Experimental Details}

\label{sec:additional-experiments}

\subsection{Simulation on Random Augmentation Graph}

\label{sec: simulation rag}
Following our setting in Section \ref{subsec:thm on random graph}, we consider a binary classification task with InfoNCE loss. We generate data from two uniform distributions on a unit ball $\sS^2$ in the $3$-dimensional space. One center is $(0,0,1)$ and another is $(0,0,-1)$. The area of both parts are 1. We take 5000 {samples} as train set and 1000 samples as test set. For the encoder class $\mathcal{F}$, a single-hidden-layer neural network with softmax activation and 256 output dimensions is adopted, which is trained by InfoNCE loss.

\subsection{The Calculation Process of ARC}
\label{exp: ARc}
Our experiments are mainly conducted on the real-world data set CIFAR-10. We use SimCLR as the training framework and ResNet18 as the backbone network. When calculating ARC, we set the number of intra-anchor augmented views $C=6$ and $k=1$. We train the network for 200 epochs and use the encoder trained with 200 epochs as the final encoder while the encoder trained with 1 epoch as the initial encoder. When we present the relationship between ARC and linear downstream accuracy trained by different augmentations, we adapt log operator on ARC.

\subsection{The Comparison between Different Unsupervised Metrics}
\label{sec:different metric}

The experiments are conducted on CIFAR-10 data set with different data augmentations including 1) RandomResizedCrop with 12 different scales of the crop, 2) ColorJitter with 10 different groups of parameters (contrast, brightness, saturation, hue), 3) composition of all augmentations used in SimCLR with 11 groups of different parameters. For computing ARC, we follow the default settings described in Appendix \ref{exp: ARc}. For rotation accuracy, we train a 4-dimension linear classifier following the fixed encoder to predicted the angles of the rotation $(0^\circ, 90^\circ,180^\circ,270^\circ)$.

{
\subsection{Analysis on the Impracticality of the Conditional Independence Assumption}
\label{sec:unpractice_of_CI}

Intuitively, as illustrated in Figure 2, a positive pair is constructed by applying different augmentations to the same natural image. As a result, the two resulting inputs are inherently dependent, making the conditional independence assumption {impractical} in real-world contrastive learning scenarios.

To further demonstrate this, we {conduct an empirical analysis to quantify the deviation from the conditional independence assumption}. Since the exact conditional probabilities $p(x, x^+ \mid y)$ and $p(x\mid y)p(x^+\mid y)$ are inaccessible, we approximate them using a pretrained encoder $f$ and the cosine similarity between features. Specifically, we use $\frac{f(x)^\top f(x^+)}{\sum\limits_{y(x')=y(x)}f(x')^\top f(x'^+)}$ to estimate $p(x,x^+|y)$ and use $\sum\limits_{x^+}\frac{f(x)^\top f(x^+)}{\sum\limits_{y(x')=y(x)}f(x')^\top f(x'^+)}$ to estimate $p(x|y)$. We then compute the ratio between the estimated joint and marginal probabilities. 

In our experiment, we use a ResNet-18 encoder pretrained on CIFAR-10 using SimCLR~\citep{simclr}. As shown in Table~\ref{tab:empirical_CI_assum}, the joint probability is significantly larger than the product of the marginals (ratio $\approx$ 3.83), indicating a notable dependence between the inputs. 

These results support the claim that the conditional independence assumption is {impractical} in realistic scenarios.

\begin{table}[!htbp]
\caption{Comparison between the estimated joint and marginal probabilities. Estimation is performed using an encoder $f$ (ResNet-18 pretrained on CIFAR-10 with SimCLR) and cosine similarity between features.}
    \centering
\begin{tabular}{@{}ccc@{}}
\toprule
$p(x,x^+|y)$ & $p(x|y)p(x^+|y)$ & ratio($\frac{p(x,x^+|y)}{p(x|y)p(x^+|y)}$) \\ \midrule
0.0023       & 0.0006           & 3.83     \\ \bottomrule
\end{tabular}
    \label{tab:empirical_CI_assum}
\end{table}
}

\newpage

\bibliography{22-1009}

\end{document}